\pdfoutput=1
\documentclass[11pt]{article}

\usepackage{acl}

\usepackage{times}
\usepackage{latexsym}

\usepackage[T1]{fontenc}
\usepackage[utf8]{inputenc}

\usepackage{microtype}

\usepackage{hyperref}
\usepackage{url}

\usepackage{subcaption}
\usepackage{multirow}
\usepackage[shortlabels]{enumitem}
\usepackage{microtype}
\usepackage{graphicx}
\usepackage{booktabs} % for professional tables
\usepackage{hyperref}
\usepackage{ifthen}
\usepackage{cuted}
\usepackage{makecell}
\usepackage{tabularx}
\usepackage{multicol}
\usepackage{wrapfig}

\usepackage{amsmath}
\usepackage{amssymb}
\usepackage{amsthm}
\usepackage{dsfont}
\usepackage{multirow}
\usepackage[shortlabels]{enumitem}

\usepackage{wasysym}
\usepackage{footmisc}
\usepackage{hyperref}

\usepackage{amsmath,amsfonts,bm}

\def\eqref#1{equation~\ref{#1}}
\def\1{\bm{1}}

\DeclareMathAlphabet{\mathsfit}{\encodingdefault}{\sfdefault}{m}{sl}
\SetMathAlphabet{\mathsfit}{bold}{\encodingdefault}{\sfdefault}{bx}{n}

\usepackage[ruled,boxed,lined]{algorithm2e} % linesnumbered
\let\oldnl\nl% Store \nl in \oldnl
\newcommand{\nonl}{\renewcommand{\nl}{\let\nl\oldnl}}% Remove line number for one line

\title{QuALITY: Question Answering with Long Input Texts, Yes!}

\author{Richard Yuanzhe Pang$^{*}$~~~Alicia Parrish$^{*}$~~~Nitish Joshi$^{*}$~~~Nikita Nangia~~~Jason Phang\\ 
\textbf{Angelica Chen~~~Vishakh Padmakumar~~~Johnny Ma~~~Jana Thompson~~~He He}\\
\textbf{Samuel R. Bowman}\\
New York University\\
{\tt \{yzpang,alicia.v.parrish\}@nyu.edu}}

\begin{document}
\maketitle

{
\let\thefootnote\relax\footnote{$^{*}$~Equal contribution.}
}

\begin{abstract}
To enable building and testing models on long-document comprehension, 
we introduce QuALITY, a multiple-choice QA dataset with context passages in English that have an average length of about 5,000 tokens, much longer than typical current models can process. 
Unlike in prior work with passages, our questions are written and validated by contributors who have read the entire passage, rather than relying on summaries or excerpts. In addition, only half of the questions are answerable by annotators working under tight time constraints, indicating that skimming and simple search are not enough to consistently perform well. 
Our baseline models perform poorly on this task (55.4\%) and significantly lag behind human performance (93.5\%). 
\end{abstract}

\section{Introduction}
\label{sec:intro}

\begin{figure}[ht!]
     \centering
         \includegraphics[width=1.00\columnwidth]{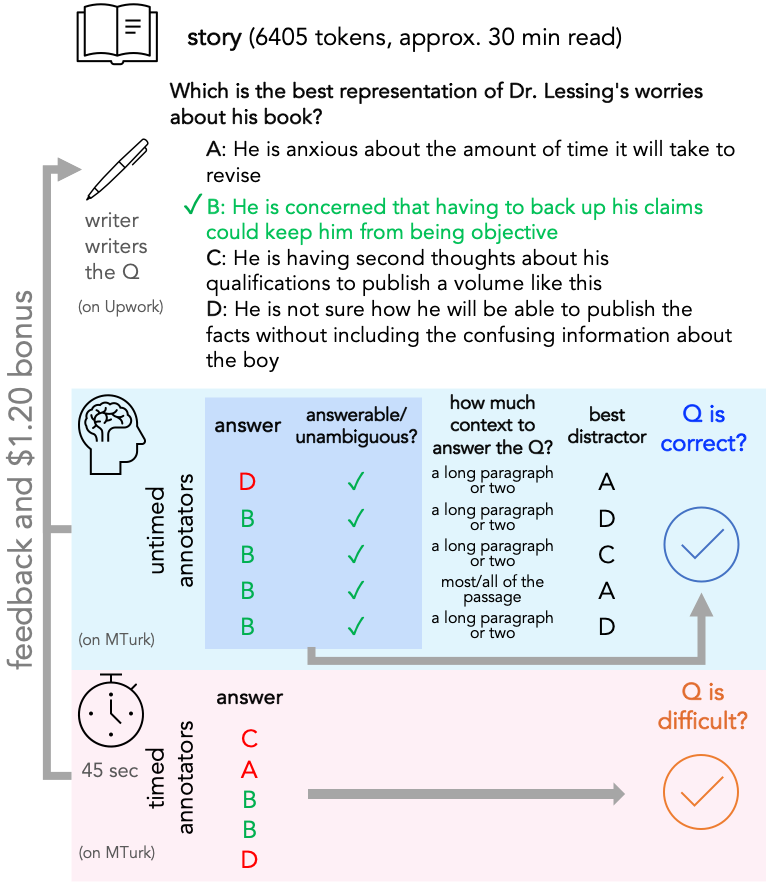}
    \caption{The crowdsourcing pipeline with an example. One writer reads the passage and writes 10 questions. Each question is validated by three or five annotators who read the full article, plus five more who have only 45 seconds per question. Writers receive feedback from both validations between writing batches. If a majority of timed annotators get the question wrong, but the untimed annotators get it right, we classify the example as \textsc{hard} and give the writer a bonus.}
    \label{fig:flowchart}
    \vspace{-0.3cm}
\end{figure}

Most of the best models for natural language understanding are restricted to processing only a few hundred words of text at a time, preventing them from solving tasks that require a holistic understanding of an entire passage. Moving past this limitation would open up new applications in areas like news comprehension, summarization, or applied question answering.
We think that new benchmark datasets will help us do this.
Most existing datasets \citep{rajpurkar-etal-2018-know,fan-etal-2019-eli5,lelkes2021quiz} use shorter contexts that humans can read within a few minutes. While there are open-domain QA datasets that require longer contexts \citep{joshi-etal-2017-triviaqa,zhu2021retrieving}, finding a short excerpt that answers the questions at hand often suffices; however, long-document QA requires understanding a long context as a whole to correctly answer questions. 

% putting the table here so it appears on the second page
\begin{table*}[th]
\setlength{\tabcolsep}{3.3pt}
\centering
\small
\begin{tabular}{p{10ex}lp{21ex}p{63ex}r}
    \toprule
    Source & Dif. & Question & Answer Options & Label \\
    \midrule
    \multirow{2}{10ex}{Gutenberg} & Hard & Why was the Volpla vocabulary limited when the narrator took a few into the valley?
 & (a) They had not been alive long enough to learn enough English to communicate well (b) They were encountering concepts that were unfamiliar from the lab environment (c) They are not smart enough to have a fully developed language, no matter how hard they try (d) They were confusing their own language with English, having trouble keeping the languages separate & b \\ 
    {} & Easy & What is Russell's greatest fear? & (a) Being disappointed (b) Losing his mind (c) Being lost and alone (d) Living forever & c \\
    \midrule
    \multirow{2}{10ex}{Slate} & Hard & Which is NOT a reason why the narrator is concerned with the antichrist? & (a) Evangelical Christians are preaching that the end of the world is coming soon. (b) He is concerned that Christians will become violent toward Jews. (c) He thinks his life will be more important and influential than the average person. (d) He is conducting research for his dissertation. & d \\ 
    & Easy & Why does the author tell a story about his vehicle? & (a) To talk about how fast he drives (b) To make a point about what has the most impact on the economy (c) To talk about safe driving speeds (d) To make a point about how many different things impact the unemployment rate & b \\
    \midrule
    \multirow{2}{10ex}{Misc.} & Hard & How does Sara feel about the Chevrolet ad? & (a) She thinks it's a final chance to bond with her father (b) She is sorry she did not watch the whole ad before she reacted to it (c) She is upset at the glorification of the military (d) he is frustrated that it tokenized a Mexican family & b \\ 
    {} & Easy & Why did Birmingham build over the Victorian era relics? & (a) To create space for a Maglev train (b) To erase their history (c) They were running out of room (d) To make technological progress & d \\
    \bottomrule
\end{tabular}
\caption{Representative examples randomly selected from the training and dev sets in QuALITY. 
}
\label{tab:full_examples}
\end{table*}

NarrativeQA \citep{kocisky-etal-2018-narrativeqa} is the most established existing long-text benchmark for language understanding. It is a free-text-response QA dataset built around movie scripts and books, with an average of about 63k tokens of input per question. The authors creatively use \textit{summaries} of the texts as the basis for their questions to make data collection relatively efficient. 
This protocol leads to short answers (avg. 4.7 tokens), and few questions require complex explanation-based reasoning: >60\% are what/who questions and <10\% are why/reason questions.  
Further, the sources are usually famous, such that they are analyzed and discussed widely in the training data used by large language models. 
Additionally, the generation-based format comes with the hurdle of determining how to fairly assess accuracy, as metrics like BLEU, ROUGE, or BERTScore may not accurately convey the quality of generations \citep{wang-etal-2020-asking, durmus-etal-2020-feqa}. To ease the burden of evaluation, we opt for a multiple-choice format to evaluate a model's long-document understanding ability.

We introduce our dataset QuALITY, Question Answering with Long Input Text, Yes!\footnote{The data is available, along with links to code and additional resources, at \url{https://github.com/nyu-mll/quality}. Results on QuALITY will be collected on a dedicated leaderboard at \url{https://nyu-mll.github.io/quality/} and through the multitask SCROLLS benchmark \citep{shaham2022scrolls}.\label{footnote:leaderboard}} 
This is a multiple-choice QA dataset that uses English source articles of 2k--8k tokens.\footnote{1.5k--6k words, not counting punctuation.} 
We collect this dataset using a creative crowdsourcing pipeline that ensures the examples have unambiguous answers but are still challenging. 
We instruct example writers to carefully read the full source article before writing questions, and to then write questions that are unambiguous and require consolidating information from multiple parts of the text. Then, to ensure our questions require readers to understand the larger context from the passage, in addition to running standard validation where annotators read the text and answer the questions, we also run speed validation (\S\ref{sec:all-validation}). In speed validation, annotators only have access to the text for 45 seconds, so they can only skim or search for phrases to answer the question. If a question is unanswerable in this setting but unambiguous and answerable in the standard untimed setting, we use it as a signal for question difficulty.
This crowdsourcing process is slow and expensive (\$9.10/question),\footnote{This includes the cost of question writing and validation for questions that were discarded after validation.} but we successfully collect a challenging, high-quality, long-document multiple-choice QA dataset.
QuALITY has 6,737 questions in total, of which 3,360 questions are in the difficult subset, QuALITY-\textsc{hard}.
Table \ref{tab:full_examples} shows representative \textsc{easy} and \textsc{hard} examples from different types of source texts.

We test the Longformer (including the LED variant), RoBERTa, DeBERTaV3, and T5 models, using as much of the full source text as possible. In particular, for models whose context lengths are much shorter than the average article length, we test two-step systems with an extraction step that passes shorter contexts to the QA model. For text extraction, we use ROUGE-1 recall, fastText, or DPR based matching with the questions. The best model performance is achieved by DeBERTaV3-large with DPR-based extraction, with an accuracy of 55.4\%. The best model's accuracy on QuALITY-\textsc{hard} is 46.7\%. Model accuracy is far below human accuracy on QuALITY, where human accuracy is 93.5\% on the full dataset and 89.1\% on QuALITY-\textsc{hard}.

\section{Data Collection}

\subsection{Overview}
\label{sec:overview}

\paragraph{Sources}

In order to create a dataset that is both broadly usable and meets the goal of containing long input texts, we use only sources that are licensed under CC-BY (or more permissive licenses) and contain articles of at least 2k tokens that are likely to allow for complex questions.
We ultimately use Project Gutenberg fiction stories (mostly science fiction),\footnote{ \url{http://www.gutenberg.org}} Slate magazine articles from the Open American National
Corpus \citep{fillmore1998american,ide-suderman-2004-american}, and other nonfiction articles taken from The Long+Short,\footnote{ \url{http://thelongandshort.org}} Freesouls,\footnote{\url{http://freesouls.cc}} and the book Open Access \citep{openaccess}. Table \ref{tab:data-splits} shows how many articles and questions come from each.
Most of the Gutenberg texts are from the 1950s--1970s, while the other texts are mostly from the 1990s and after.

Texts are provided with the original HTML tags indicating paragraph breaks and basic formatting (e.g., italics), and it is in this format, with images removed, that we present the texts to our writers and annotators. In our dataset release, we also include a version of each file with this information stripped away, as current models, including our baselines, are not trained to consume these tags.

We set a maximum length for the texts at 6k words using word-level tokenization without counting HTML tags.\footnote{With spaCy tokenization, the maximum number of tokens is larger (Figure \ref{fig:lengths}).} 
For around 40\% of the Gutenberg articles, the full text data is much longer; in these cases, we truncate the texts and manually check to make sure the truncation happens at a reasonable location (i.e., not in the middle of a paragraph).

\paragraph{Stages of Data Collection}

We collect data over several rounds to provide writers with feedback throughout the process.
We iterate through the following pipeline each round: (i) we assign writers a set of passages, and they write 10 questions for each (\S \ref{sec:writing}), (ii) annotators complete speed validation (\S \ref{sec:speed-val}), (iii) annotators complete untimed validation (\S \ref{sec:untimed_val}), and (iv) we award writers bonuses and send feedback based on the annotations.

\subsection{Question Writing}
\label{sec:writing}

We hire 22 writers---most with degrees or professional experience in literature or teaching---from the freelancing platform Upwork and design a multi-part incentive structure to encourage difficult yet answerable questions. 
Details about hiring and writer qualifications are in Appendix \ref{app:writer_recruitment}.

\paragraph{The Writing Task}

We design a feedback and incentive structure to encourage writers towards questions that are answerable, unambiguous, and difficult. 
Writers construct examples over multiple rounds, and they receive (i) detailed feedback based on the two validation tasks and (ii) bonuses based on how many of their questions met our criteria for \textsc{hard} questions.
Each writer constructs 10 questions with four answer options for a given passage, and they complete 6--30 such passages each round. Each passage is assigned to two writers, so there are 20 questions for each passage before filtering. 
Writers earn an average rate of \$21.05/hr, after bonuses.
Details about this process and the timeline are in Appendix \ref{app:writing_task}.

\subsection{Data Validation}
\label{sec:all-validation}

We use two validation tasks to evaluate if (i) the questions are difficult by testing if they are answerable under strict time constraints (speed validation) and (ii) the questions have a single correct answer (untimed validation).
We recruit 45 annotators via Amazon Mechanical Turk (MTurk); details on the qualification process are in Appendix \ref{app:annotator-recruitment}.

\subsubsection{Speed Validation}
\label{sec:speed-val}

We want to ensure that the questions require understanding of the full text to answer correctly. 
If a person can quickly identify the answer to a question, such as through skimming or ctrl-F-style in-browser search, then the question does not require broader understanding of the passage, and a model is likely to be able to identify the correct answer via extractive methods. 
More precisely, we aim to collect questions for which annotators, in the aggregate, are unable to select the correct answer under strict time constraints, and we construct a speed validation task to test this. Questions that pass this bar make up the \textsc{hard} subset of QuALITY.

To our knowledge, this is a novel data collection method---it is inspired by adversarial collection methods \cite{nie-etal-2020-adversarial,bartolo-etal-2020-beat} as a way of collecting more challenging data. 
As model performance on our dataset is very low, a true adversarial design would not be practical because model behavior would not provide enough signal to the crowdworkers, and we would risk limiting the usefulness of QuALITY as a test set for a full range of models \cite{bowman-dahl-2021-will}. 
Thus, we design this task in a way that writers can reason about what would be difficult for \textit{another human} as opposed to a model, and we award bonuses based on that metric. 

\paragraph{Procedure}

We collect five annotations per question; within each task, questions appear one at a time to ensure the time limit is consistent for each question.  
The worker first reads the question and the four answer options without access to the passage. Then they press a button to reveal the passage, and they have 40 seconds to skim or search for keywords (e.g., with ctrl+F) to determine the correct answer.
After the timer runs out, the passage disappears, and they have 5 more seconds to select an answer. Appendix \ref{app:ui} shows the user interface. 
% set up -- 10 questions per trial, 9 test and 1 catch, passages randomized 

Each task consists of 10 questions from different passages, and the order of the answer options is randomized.
Within each task, there are nine questions written by the Upwork writers and one question written by the authors as a catch question. 
We pay workers \$2.25 per task and award a bonus of \$0.20 for each correct answer. 
On average, workers earn a bonus of \$1.03 per task, and we estimate based on workers' survey responses that each task takes 11-12 minutes, for an effective rate of just over \$17/hr.
We use the catch questions to track annotator performance and ensure that all workers are performing well above chance on these examples, indicating that they are consistently making a faithful effort to find the answer in the text (see Appendix \ref{app:speed_val} for additional details on the task, catch questions, and annotator performance).

\subsubsection{Untimed Validation}
\label{sec:untimed_val}

To ensure all questions in QuALITY are correct and unambiguous, we conduct a validation task without a time limit, but with strong incentives towards accuracy. 
We collect three annotations for each example in the training set, and five annotations for each example in the dev and test sets.

\paragraph{Procedure}
% set up -- 20 questions per passage, randomized
Each task consists of one passage with all 20 questions created by the writers.
Each of the 20 reading comprehension questions has three evaluation questions immediately below it.
We instruct workers to first read the passage carefully and then answer all the questions.
% compensation -- base pay \& bonuses
Each task pays \$6.50, with a \$0.50 bonus for each question in which both the reading comprehension question and evaluation question 1 (see below) agree with the majority vote label.\footnote{Bonuses were \$0.40 in round 1 and based only on reading comprehension questions. We updated both following our evaluation of the results and worker feedback.} 
We estimate based on survey responses that workers spend about 50--60 minutes on this task; the average bonus rate is \$8.13 per task, for an average rate of \$15.96/hr.

\begin{figure*}[t!]
    \centering
    \begin{subfigure}[t]{0.325\textwidth}
        \centering
        \includegraphics[scale=0.33]{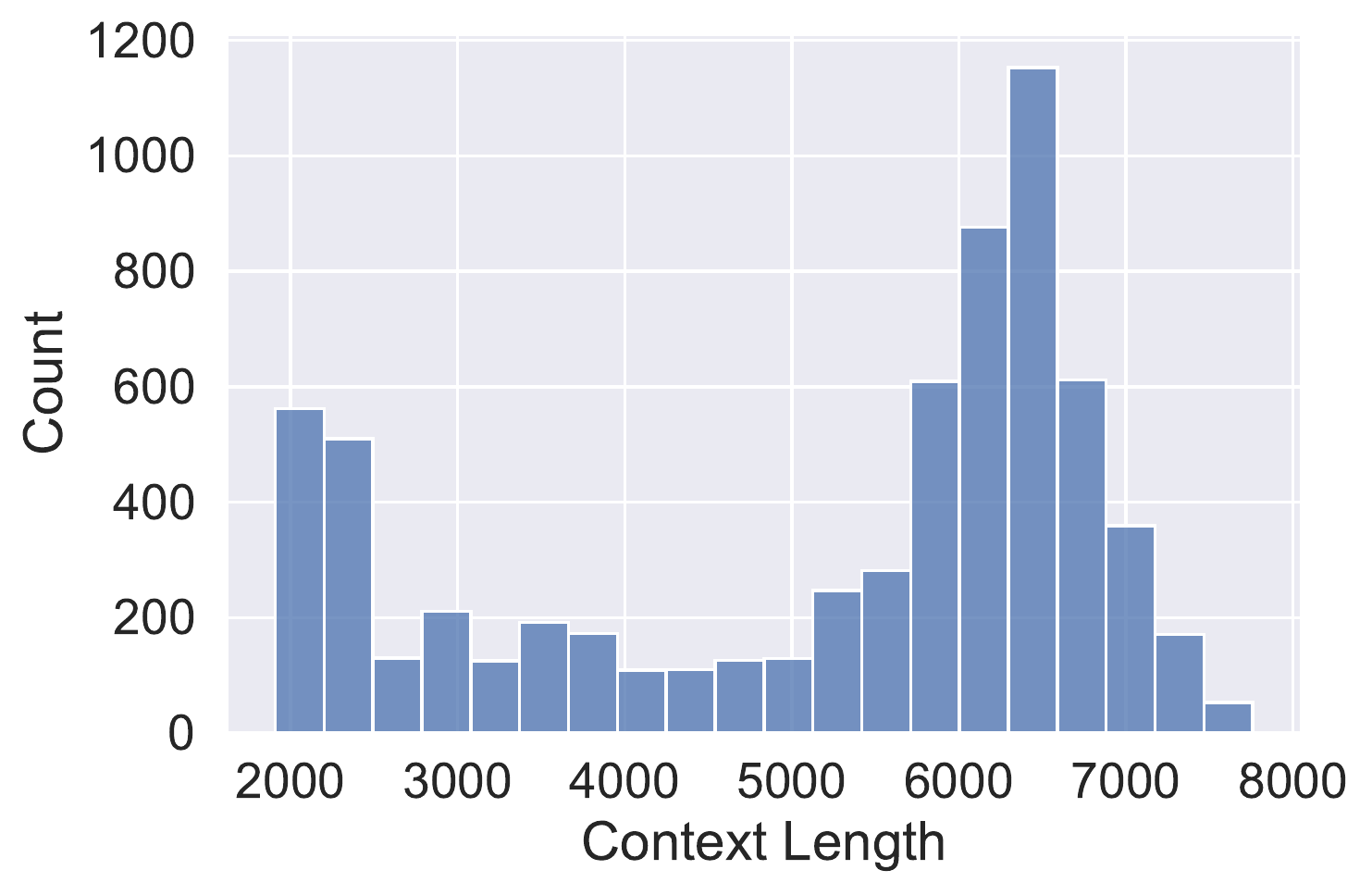}
        % \caption{}
        \label{fig:context_length}
    \end{subfigure}
    \begin{subfigure}[t]{0.325\textwidth}
        \centering
        \includegraphics[scale=0.33]{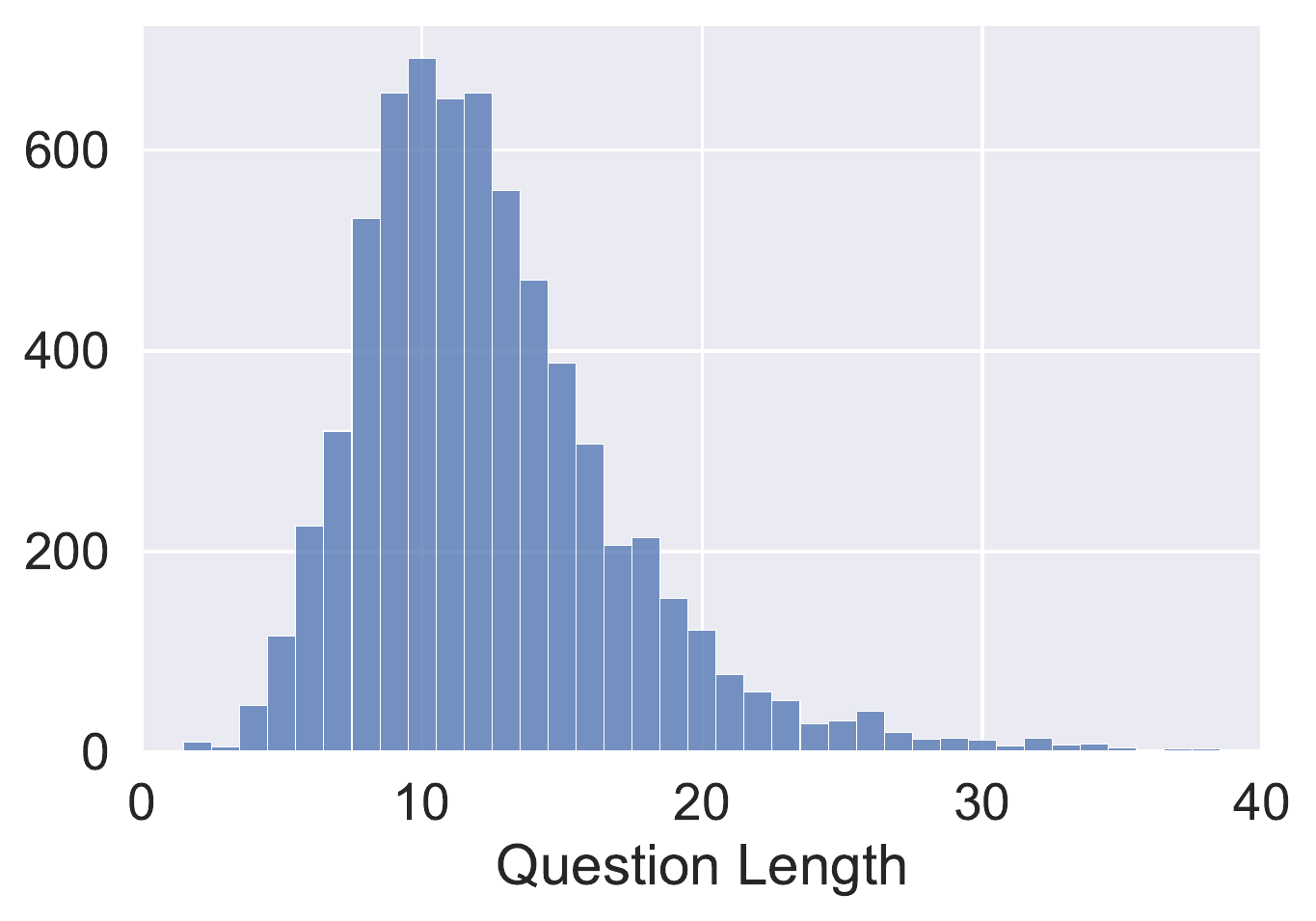}
        % \caption{}
        \label{fig:question_length}
    \end{subfigure}
    \begin{subfigure}[t]{0.325\textwidth}
        \centering
        \includegraphics[scale=0.33]{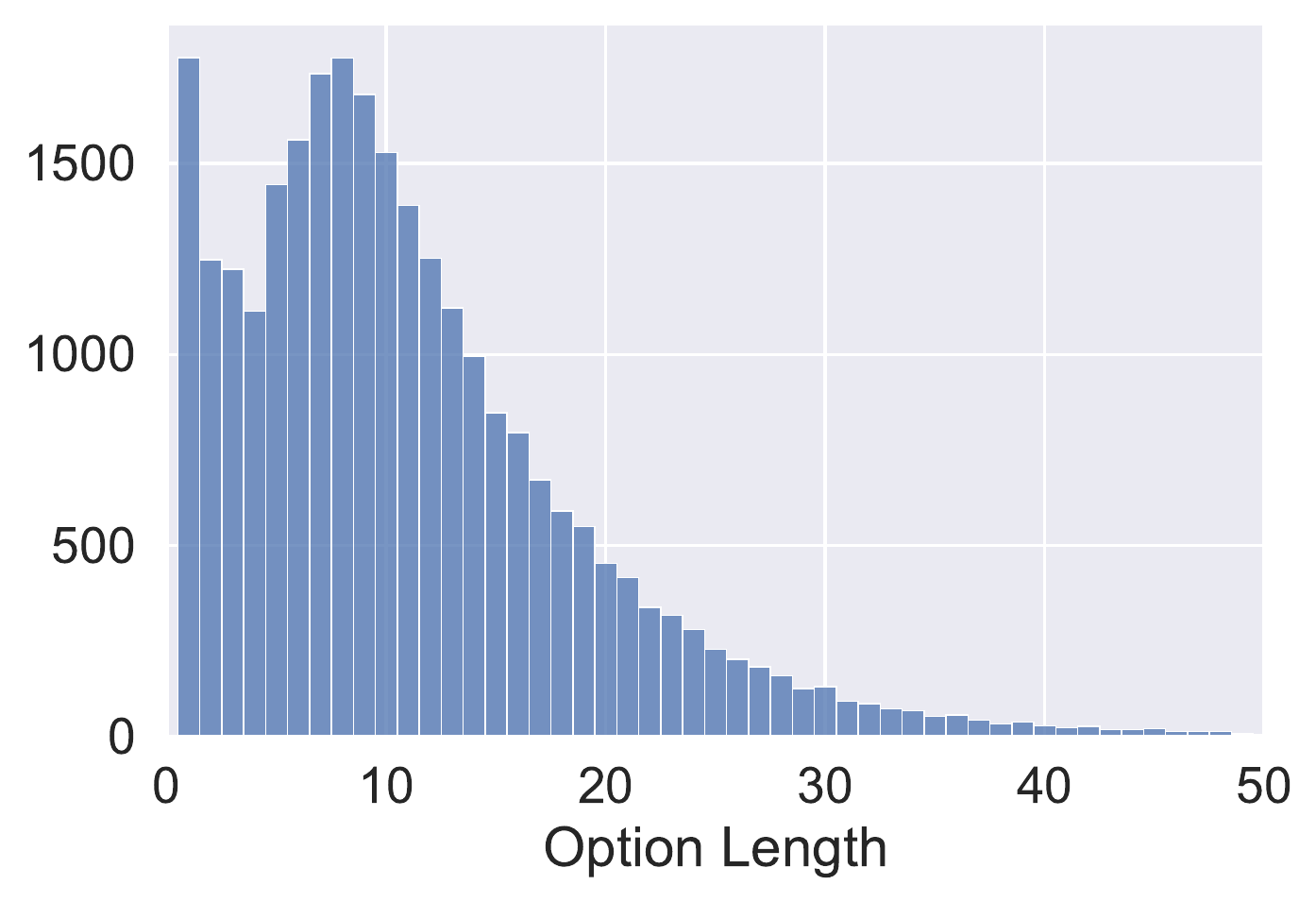}
        % \caption{}
        \label{fig:option_length}
    \end{subfigure}
    \caption{Article length, question length, and option length in QuALITY. The average length of an article, question, and option is 5,159 tokens, 12.5 tokens, and 11.2 tokens, respectively. The maximum length of an article, question, and option is 7,759 tokens, 103 tokens, and 75 tokens, respectively. The histograms are truncated to only keep the visible mass.}
    \label{fig:lengths}
\end{figure*}

\paragraph{Evaluation Questions} 

We ask the three evaluation questions in Table \ref{tab:evalquestions} immediately following each reading comprehension question to assess question quality.
Q1 is used to determine inclusion into the final dataset, as we exclude any questions for which the majority of annotators marked that the question was either ambiguous or unanswerable.
Q2 and Q3 are used for feedback to the writers.

\begin{table}[t]
\setlength{\tabcolsep}{3.0pt}
\centering
\resizebox{\linewidth}{!}{%
\begin{tabular}{p{19ex}p{34ex}}
    \toprule
    Question&Answer Options\\
    \midrule
    \multirow{3}{19ex}{Q1. Is the question answerable and unambiguous?} & 
    $\ocircle$ Yes, there is a single answer choice that is the most correct. \\
    {} & $\ocircle$ No, two or more answer choices are equally correct. \\
    {} & $\ocircle$ No, it is unclear what the question is asking, or the question or answer choices are unrelated to the passage. \\
    % & \\
    \midrule
    \multirow{4}{19ex}{Q2. How much of the passage/text is needed as context to answer this question correctly?} & $\ocircle$ Only a sentence or two of context \\
    {} & $\ocircle$ At least a long paragraph or two of context \\
    {} & $\ocircle$ At least a third of the passage for context \\
    {} & $\ocircle$ Most or all of the passage for context \\
    % & \\
    \midrule
    \multirow{4}{20ex}{Q3. Which of the options that you did not select was the best ``distractor'' item?} & $\ocircle$ Option 1 \\
    {} & $\ocircle$ Option 2 \\
    {} & $\ocircle$ Option 3 \\
    {} & $\ocircle$ Option 4 \\
    \bottomrule
\end{tabular}
}
\caption{Evaluation questions asked after each reading comprehension question during untimed validation.} 
\label{tab:evalquestions}
\end{table}

We find that responses to the evaluation questions slightly differ between the \textsc{hard} and \textsc{easy} subsets.
For Q1, individual raters are less likely to rate a \textsc{hard} question as answerable and unambiguous (92.8\%) compared to an \textsc{easy} question (95.1\%). 
For Q2, in the \textsc{hard} subset, 26.1\% of the time, the question is rated as needing at least a third of the context or more (the 3rd and 4th options), compared to 21.7\% of the time in the \textsc{easy} subset. In the \textsc{hard} subset, 81.3\% of the questions are rated as needing at least a long paragraph or two of context, compared to 73.9\% in the \textsc{easy} subset.

\paragraph{Annotator Performance}
We track annotator performance throughout data collection and remove any workers whose accuracy falls below 75\% in any given round. 
Annotator agreement on the reading comprehension questions for each passage is high, with a median Krippendorff's alpha of 0.71.
Agreement on Q1 is also high, with 92.6\% individual agreement with the majority vote,  with the two `No' options collapsed for analysis (alpha values are less valuable on such a skewed question).
As Q2 and Q3 are more subjective, responses are noisy, with median alpha values of 0.12 and 0.21, respectively.
Additional details and our protocol for reannotating data are in Appendix \ref{app:untimed_val}.

\section{Dataset Information and Analysis}

\begin{table*}[ht!]
\setlength{\tabcolsep}{2.4pt}
\centering
\small
\begin{tabular}{lccccccccccccccccccc}
    \toprule
    & \multicolumn{4}{c}{Gutenberg} & & \multicolumn{4}{c}{Slate} & & \multicolumn{4}{c}{Misc} & & \multicolumn{4}{c}{All} \\
    \cline{2-5} \cline{7-10} \cline{12-15} \cline{17-20}
    \noalign{\smallskip}
    Split & \makecell[c]{Art.} & \makecell[c]{All \\ Qs}  & \makecell[c]{\textsc{hard} \\ Qs} & \makecell[c]{\% \\ \textsc{hard}} & & \makecell[c]{Art.} & \makecell[c]{All \\ Qs}  & \makecell[c]{\textsc{hard} \\ Qs} & \makecell[c]{\% \\ \textsc{hard}} & & \makecell[c]{Art.} & \makecell[c]{All \\ Qs} & \makecell[c]{\textsc{hard} \\ Qs} & \makecell[c]{\% \\ \textsc{hard}} & & \makecell[c]{Art.} & \makecell[c]{All \\ Qs} & \makecell[c]{\textsc{hard} \\ Qs} & \makecell[c]{\% \\ \textsc{hard}} \\
    \midrule
    Train & 118 &  2000 &  1056 & 52.8 & & 22 &  355 &  142 & 40.0 && 10 &  168 &  53 & 31.5 && 150 & 2523 & 1251 & 49.5 \\ 
    Dev &  86 & 1552 & 873 & 56.2  & & 19 &  351  & 149 & 42.5 & & 10  & 183  & 43 & 23.5 && 115 & 2086 & 1065 & 51.1 \\
    Test & 81  & 1486  & 828 & 55.7 & & 25 & 450  & 170 & 37.8 && 10  & 192  & 46 & 24.0 && 116 & 2128 & 1044 & 49.1\\
    \midrule
    All & 285  & 5038  & 2757 & 54.7 & & 66  & 1156  & 461 & 40.0 && 30  & 543  & 142 & 26.2 && 381 & 6737 & 3360 & 49.9 \\
    \bottomrule
\end{tabular}
\caption{Data splits within QuALITY. Items that did not pass untimed validation are excluded from this table. `Art.' shows the number of articles. `\textsc{hard} Qs' is the number of questions in QuALITY-\textsc{hard}.} %, containing questions that annotators could not correctly answer in speed validation, but did correctly answer in untimed validation.}%and `\% \textsc{hard}' are the percentages of all questions that are labeled as hard.}
\label{tab:data-splits}
\end{table*}

After aggregating the labels assigned via untimed validations with the original writer's label, we calculate the gold label via majority vote of annotators.\footnote{This gold label calculation follows MNLI \citep{williams-etal-2018-broad} but is more conservative as the writer's label is never a tie-breaking vote. % We hypothesize that writers sometimes mislabel the correct option, because many say that they usually work on a separate document before filling in our data collection UI. We analyze a random subset of the examples where the writer's label and the gold label do not match, and confirm that the assigned gold labels are correct. 
The gold label and writer's label, both provided in the dataset, differ for $\sim$4\% (274/7620) of questions.} 
We only keep questions for which (i) a majority vote label (strictly larger than $50\%$) can be assigned 
and (ii) the majority of annotators rate the questions as answerable and unambiguous. 
6,737 out of 7,620 (88.4\%) questions meet these inclusion criteria. % \footnote{We release the discarded questions corresponding to passages in the train and dev sets as part of a supplemental dataset.} 
The \textsc{hard} subset corresponds to questions that the majority of the annotators answer incorrectly in the speed validation setting, and this constitutes 49.9\% of the final dataset.

\subsection{Human Accuracy} \label{sec:human-acc}
We estimate human accuracy on QuALITY on a random sample of 20 passages (367 questions).
Each question is annotated by 3 new annotators who had not previously annotated that passage, and whose labels do not contribute to the assignment of the gold label.
We calculate the majority vote of the annotators, which yields an accuracy of 93.5\% relative to the gold label. This breaks down to 89.1\% on the \textsc{hard} subset and 97.0\% on the \textsc{easy} subset. 
% In 37.5\% of the items that human evaluators got wrong, all three predicted answers differed. 
Annotators marked 98.5\% of questions as answerable and unambiguous.

\subsection{Size and Splits}

We split the data into train/dev/test sets\footnote{The test labels are not publicly released. The test set performance can be obtained by submitting to a leaderboard discussed in footnote \ref{footnote:leaderboard}.} such that there is minimal overlap in question writers among train/dev/test sets. 
This ensures that a model will not be rewarded for overfitting to any idiosyncrasies of a single writer's style. 
Table \ref{tab:data-splits} shows the number of articles in QuALITY and \textsc{hard} questions for each of the split. 
Gutenberg sources result in the highest proportion of \textsc{hard} questions, and misc. sources result in the lowest proportion.

\subsection{Length}

% \paragraph{Article Length}

Figure \ref{fig:lengths} shows the article lengths. The two peaks in the histogram correspond to articles from Slate and Gutenberg. The average context length is 5,159 tokens, much longer than other existing challenging QA datasets---CosmosQA \cite{huang-etal-2019-cosmos} and RACE \cite{lai-etal-2017-race} contain an average context length of 70 and 322 tokens, respectively. 
%
% \paragraph{Question and Option Lengths} 
We plot question length and option length in Figure \ref{fig:lengths} as well. The average question length is 12.5 tokens, and the average option length is 11.2 tokens.

\subsection{Lexical Overlap}
\label{ssec:overlap}

% \begin{figure}[t!]
%         \centering
%         \includegraphics[scale=0.42]{tables_figures/fig_lexical_overlap_max.pdf}
%     \caption{Lexical overlap of the correct and incorrect options with the context. Since each question has three incorrect options, we use the option with the highest lexical overlap.} 
%     \label{fig:lexical_overlap_max}
% \end{figure}

Prior work has shown that a lot of questions in existing datasets such as SQuAD can be answered by exploiting lexical overlap of the question with the article \cite{weissenborn-etal-2017-making}. To understand how effective this heuristic is in QuALITY, we compute the lexical overlap between the options and the article in QuALITY. The lexical overlap is computed as the fraction of the tokens in the option which are present in the article. Figure \ref{fig:lexical_overlap_max} in the Appendix plots the distribution of lexical overlap for the correct options and the incorrect options---since each question has three incorrect options, we use the maximum lexical overlap among the three. Simply predicting the option with the highest lexical overlap achieves only 26.6\% accuracy, so correct options do not have a higher lexical overlap than the incorrect options, making it difficult for models to rely on this heuristic.

\subsection{Question Types}
\label{sec:question-types}

\begin{table}[th]
\setlength{\tabcolsep}{3.0pt}
\centering
\small
\begin{tabular}{lrrr}
    \toprule
    Question Type & \# \textsc{easy} & \# \textsc{hard} & \% total  \\
    \midrule
    what & 1361 & 1471 & 42.2  \\
    why & 832 & 825 & 24.6  \\
    how & 385 & 416 & 11.9  \\
    which & 253 & 244 & 7.4  \\
    who & 151 & 132 & 4.2  \\
    how + meas. & 51 & 75 & 1.9 \\
    yes/no & 53 & 55 & 1.6  \\
    where & 43 & 42 & 1.3  \\
    when & 35 & 34 & 1.0  \\
    other & 155 & 124 & 4.1  \\
    \bottomrule
\end{tabular}
\caption{Different question types in QuALITY, split by \textsc{hard} and \textsc{easy} subsets. 
% Most of the questions in the `other' category are finish-the-phrase style questions, and for the example in the table, the answer options are different ways that sentence could be completed. 
`How + meas.' collapses multiple questions with `how' plus some measurement, such as `how long' or `how many.' Examples of each question type are in Appendix Table \ref{tab:questiontypes}.
}
\label{tab:questiontypes_noexamples}
\end{table}

We analyze the proportion of question types % that occur in QuALITY 
by automatically categorizing each question based on the first question word it contains.\footnote{In cases where the question starts with an auxiliary verb, or where there is no question word but an auxiliary verb appears after a comma, we categorize the question as ``yes/no.''}
Table \ref{tab:questiontypes_noexamples} shows that QuALITY contains many questions that require complex responses about ``how'' and ``why'' an event happened in a greater proportion of cases than similar datasets such as NarrativeQA.
However, we do not observe that our measure of question difficulty varies by question type.

\subsection{Reasoning Strategies}
As a qualitative analysis, we manually annotate the reasoning strategy needed in each question and present the results in Table \ref{tab:reasoning}.
We take a random subset of 500 questions from the full dataset and manually annotate them.
Each question is annotated by two of the authors; any disagreements in categorization are resolved via discussion.
As we do not read the full passages, it is not always possible to determine the reasoning strategy, but we consider both the question and answer options in categorizing each item. 

We find that many of the questions rely on (i) reasoning about the best description, (ii) determining the correct explanation for why something happened, or (iii) the reader making an interpretation or using symbolism.
All three of these reasoning types are likely to rely on broader context from the passage, compared to questions about who did something or where something happened.
For example, the question \textit{How do you think Meredith feels about the rest of the crew?} requires a description of the character's feelings (description), and it also requires the reader to interpret the character's feelings (symbolism/interpretation) and reason about the relation between different characters (relation).
We also find that, despite questions using ``what'' being the most frequent in the question-types analysis, very few of the questions in QuALITY depend on reasoning about objects or entities.
Rather, most of these ``what'' questions ask for the description of a person or situation, or they ask for an interpretation from the reader.
Further details about this analysis, the categories used, and examples of each reasoning type are in Appendix \ref{app:reasoningtypes}.

\begin{table}[th]
\setlength{\tabcolsep}{3.0pt}
\centering
\small
\begin{tabular}{lrrr}
    \toprule
    Reasoning Type & \# \textsc{hard}/ & \# \textsc{easy}/ & \% of  \\
    {} & 251 & 249 & total \\
    \midrule
    Description & 89 & 77 & 33.2 \\
    Why/reason & 73 & 83 & 31.2 \\
    Symbolism/interpretation & 76 & 63 & 27.8 \\
    How/method & 25 & 19 & 8.8 \\
    Event & 17 & 18 & 7.0 \\
    Person & 11 & 17 & 5.6 \\
    Not/except & 13 & 6 & 3.8 \\
    Relation & 12 & 7 & 3.8 \\
    Entity & 7 & 9 & 3.2 \\
    Finish the Phrase & 3 & 12 & 3.0  \\
    Location & 5 & 7 & 2.4 \\
    Numeric & 5 & 6 & 2.2 \\
    Object & 5 & 4 & 1.8 \\
    What if & 3 & 4 & 1.4 \\
    Duration & 1 & 2 & 0.6 \\
    \bottomrule
\end{tabular}
\caption{Qualitative assessment on a random 500 example subset of QuALITY, split by difficulty, and categorizing the different kinds of things that need to be reasoned about. %These categories are inspired by those used in NarrativeQA and adapted for the reasoning types observed in our data. 
Questions can require multiple reasoning types, so values do not add up to 100\%. } %The \textsc{hard} questions are slightly more likely to have multiple reasoning types.}
\label{tab:reasoning}
\end{table}

\section{Baseline Experiments}
\label{sec:modeling}

\begin{table*}[th]
\setlength{\tabcolsep}{3.6pt}
\centering
\small
\begin{tabular}{llcccccc}
    \toprule
    %& & & \multicolumn{3}{c}{Extraction Based on Qs}  \\
    %\cline{4-6} 
    %\noalign{\smallskip}
    Training Data & Model & Full & Extr: R-1 & Extr: fastText & Extr: DPR & {Question-Only} \\ 
    \midrule
    QuALITY 
      & Longformer-base & 30.7\ /\ 29.3 & -- & -- & -- & -- \\
      % & LED-base & 25.3\ /\ 25.8 & -- & -- & -- & -- \\
      & LED-base & 25.1\ /\ 24.6 & -- & -- & -- & -- \\
      & LED-large & 24.2\ /\ 24.5 & -- & -- & -- & -- \\
      & RoBERTa-base & -- & 33.4\ /\ 30.7 & 39.7\ /\ 36.1 & 39.9\ /\ 34.0 & 36.6\ /\ 34.8 \\
      & RoBERTa-large & -- & 29.4\ /\ 28.0 & 42.7\ /\ 35.7 & 26.2\ /\ 25.1 & 26.4\ /\ 25.7 \\
      & DeBERTaV3-base & -- & 36.7\ /\ 35.7 & 38.9\ /\ 35.9 & 44.1\ /\ 38.5 & 38.2\ /\ 35.6 \\
      & DeBERTaV3-large & -- & 46.5\ /\ 39.3 & 45.5\ /\ 40.2 & 49.0\ /\ 41.2 & 39.7\ /\ 35.2 \\
      & T5-base & -- & 28.0\ /\ 28.0 & 28.9\ /\ 27.4 & 29.3\ /\ 29.1 & 30.1\ /\ 29.9 \\
    \midrule 
    \multirow{3}{2cm}{RACE \\ \ \ \ \ $\downarrow$ \\ QuALITY}
      & Longformer-base & 39.5\ /\ 35.3 & -- & -- & -- & -- \\
      & LED-base & 37.2\ /\ 33.8 & -- & -- & -- & -- \\
      & LED-large & 39.4\ /\ 35.3 & -- & -- & -- & -- \\
      & RoBERTa-base & -- & 42.1\ /\ 38.3 & 43.0\ /\ 40.1 & 44.3\ /\ 39.8 & 38.1\ /\ 37.5 \\
      & RoBERTa-large & -- & 48.0\ /\ 40.8 & 50.4\ /\ 43.7 & 51.4\ /\ 44.7 & 40.4\ /\ 37.1 \\
      & DeBERTaV3-base & -- & 46.8\ /\ 38.7 & 49.8\ /\ 43.2 & 51.2\ /\ 42.4 & 41.4\ /\ 37.9  \\
      & DeBERTaV3-large & -- & 53.8\ /\ 46.3 & 54.7\ /\ \bf{46.7} & {\bf{55.4}}\ /\ 46.1 & 43.3\ /\ 38.2 \\
      & T5-base & -- & 41.1\ /\ 40.1 & 40.8\ /\ 40.1 & 41.6\ /\ 39.8 & 36.4\ /\ 35.9 \\
      \midrule
      -- & Human Annotators & 93.5\ /\ 89.1 & -- & -- & -- & -- \\
    \bottomrule
\end{tabular}
\caption{Accuracy on the full QuALITY test set and the QuALITY-\textsc{hard} subset (formatted as full\ /\ \textsc{hard}). The ``Full'' column has results from training with the source inputs truncated to fit into memory. %, without using an extractive model to select portions of text. 
R-1 (ROUGE-1), fastText, DPR are three extraction (``Extr'') methods (\S \ref{sec:models}) used to select relevant portions of the source text. 
% Poor RoBERTa-large performance (for training on QuALITY only) is likely due to unstable training given our small training set \citep{mosbach2021on}.
}
\label{tab:main-results-noRACE}
\end{table*}

\subsection{Models}
\label{sec:models}

\paragraph{Long-Context Models} 

We experiment with the Longformer model \citep{beltagy2020longformer}, which uses a combination of sliding-window local attention and global attention to encode long inputs. 
The Longformer encoder models support up to 4,096 tokens. We test Longformer because it is likely to fit most or all of the context needed to answer the questions for the majority of examples in QuALITY.\footnote{The question and answer options are visible to models, but the article is sometimes truncated.} We also experiment with Longformer Encoder-Decoder (LED) which supports up to 16,384 encoder input tokens.\footnote{Hyperparameter details for models in this section can be found in Appendix~\ref{app:modeling}.}  %\footnote{Longformer Encoder-Decoder (LED) supports up to 16,384 encoder input tokens, but due to difficulty tuning the model, our results are still pending.} 
% The maximum token limit is still a challenge, though, and we expect models that can take longer inputs to perform better. 

\paragraph{Extractive Models}

As an alternative to feeding the whole input context into a transformer model or truncating, we also test retrieval methods to score and extract relevant sentences from the passage and feed only the selected sentences as inputs to a given model. 
We can thus use a wider range of higher-performing short-sequence transformer models, at the cost of missing some input context.

Using the question as a retrieval query, we score each sentence in the passage relative to the query. 
We then select sentences in order of descending relevance until we reach 300 words.\footnote{Punctuation is not counted toward this limit.}
We then sort the selected sentences based on the original passage order and use the concatenation as the `passage' for that example. 

We consider three scoring methods. 
First, we use ROUGE-1 recall relative to the query. 
Second, we use cosine similarity based on bag-of-words of fastText \cite{bojanowski-etal-2017-enriching} embeddings. 
Third, we use DPR \citep{karpukhin-etal-2020-dense}, a model trained for open-domain retrieval for QA.
Because DPR tackles span-based question-answering, the \textit{reader} model is unsuitable for our multiple-choice dataset.
However, we can use the \textit{retriever} model for extraction, using the separate question- and context-encoders to encode our question and context sentences to vector representations. 
We then score similarity based on the negative Euclidean ($L^2$) distance. 

After extraction, we apply standard models for multiple-choice question-answering: RoBERTa \citep{liu2019roberta} and DeBERTaV3 \citep{he2021deberta} encoder models, and the T5 \citep{raffel2020t5} encoder-decoder model. 
To establish an upper bound of how well extractive models can do, we also introduce an oracle baseline in which we apply the same extraction strategy described above, but we use the correct answer as the extraction query. 

\paragraph{Question-Only Baselines}

To test for dataset artifacts, we consider a baseline where we only give the models the questions and answer options, leaving out the passage.  % We test this question-only baseline for each model type.

\paragraph{Supplementary Training Data} 

To supplement the training examples in QuALITY, we incorporate additional training examples from the RACE task dataset \citep{lai-etal-2017-race}. 
Like QuALITY, RACE is a passage-based, four-way multiple-choice question-answering dataset. Although the passages are much shorter (321.9 words on average), the training set is large ($\sim$88k questions), so we can expect reasonable knowledge transfer from RACE to QuALITY. 
We use the full RACE dataset, including both middle-school and high-school questions, for our intermediate training.

We consider three fine-tuning formats: (1) fine-tuning on QuALITY data, (2) fine-tuning on RACE and zero-shot evaluating on QuALITY, and (3) applying intermediate training \citep{phang2018stilts,pruksachatkun-etal-2020-intermediate} by first fine-tuning on RACE and then fine-tuning on QuALITY.

\subsection{Results and Analysis}
\label{sec:results-analysis}

Table \ref{tab:main-results-noRACE} shows model performance on the test set. The results on the development set and additional results from training just on RACE are in Appendix \ref{app:results}. 
All results in Table \ref{tab:main-results-noRACE} fall well below human performance. There is a gap of 38.1 points between our current best-performing model (DeBERTaV3-large trained on RACE$\rightarrow$QuALITY, using DPR-based extraction) and human performance on the full test set. On QuALITY-\textsc{hard}, this gap increases to 42.4 points.

Comparing models using different training data, we see that the RACE$\rightarrow$QuALITY results outperform RACE results in most cases (Table \ref{tab:main-results}). Fine-tuning on QuALITY contributes to a small performance gain. Both RACE and RACE$\rightarrow$QuALITY significantly outperform the QuALITY only results, likely because of the small size of the QuALITY training set, though this suggests that knowledge transfer from RACE is useful. 

In terms of extraction strategies, DPR-based extraction almost always produces the best result. In terms of models, DeBERTaV3-large consistently performs best. 
Compared to the RoBERTa and DeBERTa models fined-tuned on short contexts, the Longformer and LED models appear to struggle to learn the task from the long inputs, underperforming even the RoBERTa-base extraction-based models. 
We speculate that a combination of more long-context training data and better long-context models may improve performance beyond the extraction-based models.
As with other models, intermediate training on RACE improves performance on QuALITY.

\paragraph{Question-Only Baselines}

The best-performing question-only baseline is DeBERTaV3-large using the RACE$\rightarrow$QuALITY setting for training, achieving an accuracy of 43.3\%. The corresponding performance is only 12.1 percentage points lower than the DeBERTaV3-large's performance with text excerpts from DPR. 
This small margin of improvement may indicate that current models are not effectively using the input contexts.

\paragraph{QuALITY-\textsc{hard}}

Model performance is always lower on QuALITY-\textsc{hard} than on the full test-set, even on the question-only baselines. This suggests that speed-validation filtering yields more challenging questions for human annotators \textit{and} models. 

\paragraph{Extraction by Oracle Answer}

We show in Appendix \ref{app:results}, Table \ref{tab:oracle-results} the results of the oracle-answer-based extraction on the development set. Compared to Table \ref{tab:dev-results}, using the oracle answers for extraction improves performance significantly (topping out at 78.3\%), but is still below human performance by 15 points.  % Thus, even in the unrealistic scenario of an oracle extractor with models trained on oracle-extracted sentences, the extraction-based models still underperform humans. 
This demonstrates that extracting relevant excerpts alone is insufficient to solve QuALITY questions, and that QuALITY questions require reasoning over the full passage.

\section{Related Work}
\label{sec:related}

\citet{rogers2021qa} survey the QA dataset explosion of recent years and the many formats and types of QA datasets. %The QA datasets closest to our work either have long contexts or require consolidating information from multiple parts of the source text through multi-hop reasoning.
TriviaQA \citep{joshi-etal-2017-triviaqa} and SearchQA \citep{Dunn2017SearchQAAN} contain questions with more than one document as the context, but since the supporting documents are collected after writing the question-answer pairs, most questions can be answered after retrieving a short context. HotpotQA \citep{yang-etal-2018-hotpotqa}, QAngaroo \citep{welbl-etal-2018-constructing}, and ComplexWebQuestions \citep{talmor-berant-2018-web} are constructed to have more challenging questions which require multi-hop reasoning across multiple paragraphs. However, there has been recent work \citep{jiang-bansal-2019-avoiding, min-etal-2019-compositional} showing that these datasets contain reasoning shortcuts and a large fraction of the questions can be answered with single-hop reasoning. 

NarrativeQA \citep{kocisky-etal-2018-narrativeqa}, the most similar work to ours, uses entire Gutenberg books and film scripts as contexts, with an average length of 60k tokens. The authors creatively make data-collection tractable by using Wikipedia summaries for the books as context when crowdsourcing questions.  
Unlike QuALITY, NarrativeQA is a free-form generation-based task. 
While there are many existing multiple-choice QA datasets  \citep{richardson-etal-2013-mctest, Hill2016TheGP, lai-etal-2017-race, Bajgar2016EmbracingDA, huang-etal-2019-cosmos}, they use much shorter contexts (<500 tokens) than our dataset.

A primary challenge of building a long-document QA dataset like QuALITY or NarrativeQA is building a tractable crowdsourcing pipeline that enables collecting high-quality examples. \citet{roit-etal-2020-controlled} collect a challenging QA-SRL dataset by carefully hiring and training crowdworkers, with a strict qualification followed by two hours of training with extensive feedback. \citet{nangia-etal-2021-ingredients} compare crowdsourcing methods for collecting high-quality QA data and find that a long training process with iterative feedback and qualifications is an effective strategy.

\section{Conclusion}
\label{sec:conclusion}

We introduce the long-document QA dataset QuALITY. This dataset was crowdsourced and validated by humans to ensure that the questions are answerable, unambiguous, and challenging. The QuALITY-\textsc{hard} subset, comprising half the dataset, consists of questions that are unanswerable by annotators working under tight time constraints, helping ensure that skimming and simple search do not yield high performance. 

We find that our baseline models significantly lag behind human performance on QuALITY, with a 38.1 percentage point gap between human annotators and the best performing model. The gap is even wider on QuALITY-\textsc{hard}, at 42.3 points.
We hope that research that aims at this gap will contribute to expanding the scope of texts on which effective NLU systems can be applied.

\section*{Ethical Considerations}

Both the authors of our source texts and the authors of our questions are based primarily in the US, and represent a relatively privileged, educated population. A system that performs well on our dataset is, thus, only demonstrating its effectiveness on mainstream US English, and should not be presumed to be effective on text in other languages or language varieties.

\section*{Author Contributions}

\begin{itemize}[noitemsep]
    \item Locating appropriate passage sources: AP, JM, VP, AC, NN, NJ, JT
    \item Preprocessing passages: AC, NN, NJ, JP
    \item Data collection protocol design: NN, AP, RP, SB, HH
    \item Data collection user interface: RP, JT
    \item Data collection backend infrastructure: AC
    \item Data collection management and postprocessing: RP, AP, NJ
    \item Writing catch questions: VP, JM, JT, AP, NN, NJ, RP
    \item Data analysis: AP, NJ, RP, VP
    \item Modeling: JP, AC
    \item Writing: AP, RP, NN, NJ, JP, HH, SB
    \item Project management: RP, AP
    \item Advising: SB
\end{itemize}

\section*{Acknowledgements}

This project has benefited from financial support to SB by Eric and Wendy Schmidt (made by recommendation of the Schmidt Futures program), Samsung Research (under the project \textit{Improving Deep Learning using Latent Structure}), Samsung Advanced Institute of Technology (under the project \textit{Next Generation Deep Learning: From Pattern Recognition to AI}), and Apple. This material is based upon work supported by the National Science Foundation under Grant Nos. 1922658 and 2046556. Any opinions, findings, and conclusions or recommendations expressed in this material are those of the authors and do not necessarily reflect the views of the National Science Foundation. 

We thank Jon Ander Campos, Alex Wang, Saku Sugawara, Omer Levy, and Chen Zhao for valuable discussion. We thank the anonymous reviewers for useful feedback. Finally, we thank the writers who wrote our source texts (credited in the data itself) and the writers who wrote our questions:
Megan Barbee, Bridget Barrett, Kourtney Bradley, Kyle J. Brown, Alicia Chatten, Christine D., Leah Dorschner-Karim, Bobbie Dunn, Charisse Hake, Javier Hernandez, Molly Montgomery, Carilee Moran, Tracy M. Snyder, Lorna Stevenson, Isaiah Swanson, Kyla Thiel, Lisa V., Ryan Warrick, Julia Williamson, and others who chose to remain anonymous.

\bibliography{anthology,custom}

\begin{thebibliography}{40}
\expandafter\ifx\csname natexlab\endcsname\relax\def\natexlab#1{#1}\fi

\bibitem[{Bajgar et~al.(2016)Bajgar, Kadlec, and
  Kleindienst}]{Bajgar2016EmbracingDA}
Ondrej Bajgar, Rudolf Kadlec, and Jan Kleindienst. 2016.
\newblock Embracing data abundance: Booktest dataset for reading comprehension.
\newblock \emph{arXiv preprint arXiv:1610.00956}.

\bibitem[{Bartolo et~al.(2020)Bartolo, Roberts, Welbl, Riedel, and
  Stenetorp}]{bartolo-etal-2020-beat}
Max Bartolo, Alastair Roberts, Johannes Welbl, Sebastian Riedel, and Pontus
  Stenetorp. 2020.
\newblock \href {https://doi.org/10.1162/tacl_a_00338} {Beat the {AI}:
  Investigating adversarial human annotation for reading comprehension}.
\newblock \emph{Transactions of the Association for Computational Linguistics},
  8:662--678.

\bibitem[{Beltagy et~al.(2020)Beltagy, Peters, and
  Cohan}]{beltagy2020longformer}
Iz~Beltagy, Matthew~E Peters, and Arman Cohan. 2020.
\newblock Longformer: The long-document transformer.
\newblock \emph{arXiv preprint arXiv:2004.05150}.

\bibitem[{Bojanowski et~al.(2017)Bojanowski, Grave, Joulin, and
  Mikolov}]{bojanowski-etal-2017-enriching}
Piotr Bojanowski, Edouard Grave, Armand Joulin, and Tomas Mikolov. 2017.
\newblock \href {https://doi.org/10.1162/tacl_a_00051} {Enriching word vectors
  with subword information}.
\newblock \emph{Transactions of the Association for Computational Linguistics},
  5:135--146.

\bibitem[{Bowman and Dahl(2021)}]{bowman-dahl-2021-will}
Samuel~R. Bowman and George Dahl. 2021.
\newblock \href {https://doi.org/10.18653/v1/2021.naacl-main.385} {What will it
  take to fix benchmarking in natural language understanding?}
\newblock In \emph{Proceedings of the 2021 Conference of the North American
  Chapter of the Association for Computational Linguistics: Human Language
  Technologies}, pages 4843--4855, Online. Association for Computational
  Linguistics.

\bibitem[{Dunn et~al.(2017)Dunn, Sagun, Higgins, Guney, Cirik, and
  Cho}]{Dunn2017SearchQAAN}
Matthew Dunn, Levent Sagun, Mike Higgins, V~Ugur Guney, Volkan Cirik, and
  Kyunghyun Cho. 2017.
\newblock {SearchQA}: A new {Q\&A} dataset augmented with context from a search
  engine.
\newblock \emph{arXiv preprint arXiv:1704.05179}.

\bibitem[{Durmus et~al.(2020)Durmus, He, and Diab}]{durmus-etal-2020-feqa}
Esin Durmus, He~He, and Mona Diab. 2020.
\newblock \href {https://doi.org/10.18653/v1/2020.acl-main.454} {{FEQA}: A
  question answering evaluation framework for faithfulness assessment in
  abstractive summarization}.
\newblock In \emph{Proceedings of the 58th Annual Meeting of the Association
  for Computational Linguistics}, pages 5055--5070, Online. Association for
  Computational Linguistics.

\bibitem[{Fan et~al.(2019)Fan, Jernite, Perez, Grangier, Weston, and
  Auli}]{fan-etal-2019-eli5}
Angela Fan, Yacine Jernite, Ethan Perez, David Grangier, Jason Weston, and
  Michael Auli. 2019.
\newblock \href {https://doi.org/10.18653/v1/P19-1346} {{ELI}5: Long form
  question answering}.
\newblock In \emph{Proceedings of the 57th Annual Meeting of the Association
  for Computational Linguistics}, pages 3558--3567, Florence, Italy.
  Association for Computational Linguistics.

\bibitem[{Fillmore et~al.(1998)Fillmore, Ide, Jurafsky, and
  Macleod}]{fillmore1998american}
Charles Fillmore, Nancy Ide, Daniel Jurafsky, and Catherine Macleod. 1998.
\newblock An {A}merican national corpus: A proposal.
\newblock In \emph{Proceedings of the First Annual Conference on Language
  Resources and Evaluation}, pages 965--969. Citeseer.

\bibitem[{He et~al.(2021)He, Liu, Gao, and Chen}]{he2021deberta}
Pengcheng He, Xiaodong Liu, Jianfeng Gao, and Weizhu Chen. 2021.
\newblock \href {https://openreview.net/forum?id=XPZIaotutsD} {{DeBERTa}:
  Decoding-enhanced bert with disentangled attention}.
\newblock In \emph{International Conference on Learning Representations}.

\bibitem[{Hill et~al.(2015)Hill, Bordes, Chopra, and Weston}]{Hill2016TheGP}
Felix Hill, Antoine Bordes, Sumit Chopra, and Jason Weston. 2015.
\newblock The {G}oldilocks principle: Reading children's books with explicit
  memory representations.
\newblock \emph{arXiv preprint arXiv:1511.02301}.

\bibitem[{Huang et~al.(2019)Huang, Le~Bras, Bhagavatula, and
  Choi}]{huang-etal-2019-cosmos}
Lifu Huang, Ronan Le~Bras, Chandra Bhagavatula, and Yejin Choi. 2019.
\newblock \href {https://doi.org/10.18653/v1/D19-1243} {Cosmos {QA}: Machine
  reading comprehension with contextual commonsense reasoning}.
\newblock In \emph{Proceedings of the 2019 Conference on Empirical Methods in
  Natural Language Processing and the 9th International Joint Conference on
  Natural Language Processing (EMNLP-IJCNLP)}, pages 2391--2401, Hong Kong,
  China. Association for Computational Linguistics.

\bibitem[{Ide and Suderman(2004)}]{ide-suderman-2004-american}
Nancy Ide and Keith Suderman. 2004.
\newblock \href {http://www.lrec-conf.org/proceedings/lrec2004/pdf/518.pdf}
  {The {A}merican national corpus first release}.
\newblock In \emph{Proceedings of the Fourth International Conference on
  Language Resources and Evaluation ({LREC}{'}04)}, Lisbon, Portugal. European
  Language Resources Association (ELRA).

\bibitem[{Jiang and Bansal(2019)}]{jiang-bansal-2019-avoiding}
Yichen Jiang and Mohit Bansal. 2019.
\newblock \href {https://doi.org/10.18653/v1/P19-1262} {Avoiding reasoning
  shortcuts: Adversarial evaluation, training, and model development for
  multi-hop {QA}}.
\newblock In \emph{Proceedings of the 57th Annual Meeting of the Association
  for Computational Linguistics}, pages 2726--2736, Florence, Italy.
  Association for Computational Linguistics.

\bibitem[{Joshi et~al.(2017)Joshi, Choi, Weld, and
  Zettlemoyer}]{joshi-etal-2017-triviaqa}
Mandar Joshi, Eunsol Choi, Daniel Weld, and Luke Zettlemoyer. 2017.
\newblock \href {https://doi.org/10.18653/v1/P17-1147} {{T}rivia{QA}: A large
  scale distantly supervised challenge dataset for reading comprehension}.
\newblock In \emph{Proceedings of the 55th Annual Meeting of the Association
  for Computational Linguistics (Volume 1: Long Papers)}, pages 1601--1611,
  Vancouver, Canada. Association for Computational Linguistics.

\bibitem[{Karpukhin et~al.(2020)Karpukhin, Oguz, Min, Lewis, Wu, Edunov, Chen,
  and Yih}]{karpukhin-etal-2020-dense}
Vladimir Karpukhin, Barlas Oguz, Sewon Min, Patrick Lewis, Ledell Wu, Sergey
  Edunov, Danqi Chen, and Wen-tau Yih. 2020.
\newblock \href {https://doi.org/10.18653/v1/2020.emnlp-main.550} {Dense
  passage retrieval for open-domain question answering}.
\newblock In \emph{Proceedings of the 2020 Conference on Empirical Methods in
  Natural Language Processing (EMNLP)}, pages 6769--6781, Online. Association
  for Computational Linguistics.

\bibitem[{Ko{\v{c}}isk{\'y} et~al.(2018)Ko{\v{c}}isk{\'y}, Schwarz, Blunsom,
  Dyer, Hermann, Melis, and Grefenstette}]{kocisky-etal-2018-narrativeqa}
Tom{\'a}{\v{s}} Ko{\v{c}}isk{\'y}, Jonathan Schwarz, Phil Blunsom, Chris Dyer,
  Karl~Moritz Hermann, G{\'a}bor Melis, and Edward Grefenstette. 2018.
\newblock \href {https://doi.org/10.1162/tacl_a_00023} {The {N}arrative{QA}
  reading comprehension challenge}.
\newblock \emph{Transactions of the Association for Computational Linguistics},
  6:317--328.

\bibitem[{Lai et~al.(2017)Lai, Xie, Liu, Yang, and Hovy}]{lai-etal-2017-race}
Guokun Lai, Qizhe Xie, Hanxiao Liu, Yiming Yang, and Eduard Hovy. 2017.
\newblock \href {https://doi.org/10.18653/v1/D17-1082} {{RACE}: Large-scale
  {R}e{A}ding comprehension dataset from examinations}.
\newblock In \emph{Proceedings of the 2017 Conference on Empirical Methods in
  Natural Language Processing}, pages 785--794, Copenhagen, Denmark.
  Association for Computational Linguistics.

\bibitem[{Lelkes et~al.(2021)Lelkes, Tran, and Yu}]{lelkes2021quiz}
Adam~D. Lelkes, Vinh~Q. Tran, and Cong Yu. 2021.
\newblock \href {https://doi.org/10.1145/3442381.3449892} {{Quiz-style question
  generation for news stories}}.
\newblock In \emph{Proceedings of the the Web Conference 2021}.

\bibitem[{Liu et~al.(2019)Liu, Ott, Goyal, Du, Joshi, Chen, Levy, Lewis,
  Zettlemoyer, and Stoyanov}]{liu2019roberta}
Yinhan Liu, Myle Ott, Naman Goyal, Jingfei Du, Mandar Joshi, Danqi Chen, Omer
  Levy, Mike Lewis, Luke Zettlemoyer, and Veselin Stoyanov. 2019.
\newblock {RoBERTa}: A robustly optimized {BERT} pretraining approach.
\newblock \emph{arXiv preprint arXiv:1907.11692}.

\bibitem[{Min et~al.(2019)Min, Wallace, Singh, Gardner, Hajishirzi, and
  Zettlemoyer}]{min-etal-2019-compositional}
Sewon Min, Eric Wallace, Sameer Singh, Matt Gardner, Hannaneh Hajishirzi, and
  Luke Zettlemoyer. 2019.
\newblock \href {https://doi.org/10.18653/v1/P19-1416} {Compositional questions
  do not necessitate multi-hop reasoning}.
\newblock In \emph{Proceedings of the 57th Annual Meeting of the Association
  for Computational Linguistics}, pages 4249--4257, Florence, Italy.
  Association for Computational Linguistics.

\bibitem[{Nangia et~al.(2021)Nangia, Sugawara, Trivedi, Warstadt, Vania, and
  Bowman}]{nangia-etal-2021-ingredients}
Nikita Nangia, Saku Sugawara, Harsh Trivedi, Alex Warstadt, Clara Vania, and
  Samuel~R. Bowman. 2021.
\newblock \href {https://doi.org/10.18653/v1/2021.acl-long.98} {What
  ingredients make for an effective crowdsourcing protocol for difficult {NLU}
  data collection tasks?}
\newblock In \emph{Proceedings of the 59th Annual Meeting of the Association
  for Computational Linguistics and the 11th International Joint Conference on
  Natural Language Processing (Volume 1: Long Papers)}, pages 1221--1235,
  Online. Association for Computational Linguistics.

\bibitem[{Nie et~al.(2020)Nie, Williams, Dinan, Bansal, Weston, and
  Kiela}]{nie-etal-2020-adversarial}
Yixin Nie, Adina Williams, Emily Dinan, Mohit Bansal, Jason Weston, and Douwe
  Kiela. 2020.
\newblock \href {https://doi.org/10.18653/v1/2020.acl-main.441} {Adversarial
  {NLI}: A new benchmark for natural language understanding}.
\newblock In \emph{Proceedings of the 58th Annual Meeting of the Association
  for Computational Linguistics}, pages 4885--4901, Online. Association for
  Computational Linguistics.

\bibitem[{Phang et~al.(2018)Phang, F{\'e}vry, and Bowman}]{phang2018stilts}
Jason Phang, Thibault F{\'e}vry, and Samuel~R Bowman. 2018.
\newblock Sentence encoders on {STILTs}: Supplementary training on intermediate
  labeled-data tasks.
\newblock \emph{arXiv preprint arXiv:1811.01088}.

\bibitem[{Pruksachatkun et~al.(2020)Pruksachatkun, Phang, Liu, Htut, Zhang,
  Pang, Vania, Kann, and Bowman}]{pruksachatkun-etal-2020-intermediate}
Yada Pruksachatkun, Jason Phang, Haokun Liu, Phu~Mon Htut, Xiaoyi Zhang,
  Richard~Yuanzhe Pang, Clara Vania, Katharina Kann, and Samuel~R. Bowman.
  2020.
\newblock \href {https://doi.org/10.18653/v1/2020.acl-main.467}
  {Intermediate-task transfer learning with pretrained language models: When
  and why does it work?}
\newblock In \emph{Proceedings of the 58th Annual Meeting of the Association
  for Computational Linguistics}, pages 5231--5247, Online. Association for
  Computational Linguistics.

\bibitem[{Raffel et~al.(2020)Raffel, Shazeer, Roberts, Lee, Narang, Matena,
  Zhou, Li, and Liu}]{raffel2020t5}
Colin Raffel, Noam Shazeer, Adam Roberts, Katherine Lee, Sharan Narang, Michael
  Matena, Yanqi Zhou, Wei Li, and Peter~J. Liu. 2020.
\newblock \href {http://jmlr.org/papers/v21/20-074.html} {Exploring the limits
  of transfer learning with a unified text-to-text transformer}.
\newblock \emph{Journal of Machine Learning Research}, 21(140):1--67.

\bibitem[{Rajpurkar et~al.(2018)Rajpurkar, Jia, and
  Liang}]{rajpurkar-etal-2018-know}
Pranav Rajpurkar, Robin Jia, and Percy Liang. 2018.
\newblock \href {https://doi.org/10.18653/v1/P18-2124} {Know what you don{'}t
  know: Unanswerable questions for {SQ}u{AD}}.
\newblock In \emph{Proceedings of the 56th Annual Meeting of the Association
  for Computational Linguistics (Volume 2: Short Papers)}, pages 784--789,
  Melbourne, Australia. Association for Computational Linguistics.

\bibitem[{Richardson et~al.(2013)Richardson, Burges, and
  Renshaw}]{richardson-etal-2013-mctest}
Matthew Richardson, Christopher~J.C. Burges, and Erin Renshaw. 2013.
\newblock \href {https://aclanthology.org/D13-1020} {{MCT}est: A challenge
  dataset for the open-domain machine comprehension of text}.
\newblock In \emph{Proceedings of the 2013 Conference on Empirical Methods in
  Natural Language Processing}, pages 193--203, Seattle, Washington, USA.
  Association for Computational Linguistics.

\bibitem[{Rogers et~al.(2021)Rogers, Gardner, and Augenstein}]{rogers2021qa}
Anna Rogers, Matt Gardner, and Isabelle Augenstein. 2021.
\newblock {QA} dataset explosion: {A} taxonomy of {NLP} resources for question
  answering and reading comprehension.
\newblock \emph{arXiv preprint arXiv:2107.12708}.

\bibitem[{Roit et~al.(2020)Roit, Klein, Stepanov, Mamou, Michael, Stanovsky,
  Zettlemoyer, and Dagan}]{roit-etal-2020-controlled}
Paul Roit, Ayal Klein, Daniela Stepanov, Jonathan Mamou, Julian Michael,
  Gabriel Stanovsky, Luke Zettlemoyer, and Ido Dagan. 2020.
\newblock \href {https://doi.org/10.18653/v1/2020.acl-main.626} {Controlled
  crowdsourcing for high-quality {QA}-{SRL} annotation}.
\newblock In \emph{Proceedings of the 58th Annual Meeting of the Association
  for Computational Linguistics}, pages 7008--7013, Online. Association for
  Computational Linguistics.

\bibitem[{Shaham et~al.(2022)Shaham, Segal, Ivgi, Efrat, Yoran, Haviv, Gupta,
  Xiong, Geva, Berant et~al.}]{shaham2022scrolls}
Uri Shaham, Elad Segal, Maor Ivgi, Avia Efrat, Ori Yoran, Adi Haviv, Ankit
  Gupta, Wenhan Xiong, Mor Geva, Jonathan Berant, et~al. 2022.
\newblock {SCROLLS}: Standardized comparison over long language sequences.
\newblock \emph{arXiv preprint arXiv:2201.03533}.

\bibitem[{Suber(2012)}]{openaccess}
Peter Suber. 2012.
\newblock \href {https://openaccesseks.mitpress.mit.edu/} {\emph{Open Access}}.
\newblock MIT Press.

\bibitem[{Talmor and Berant(2018)}]{talmor-berant-2018-web}
Alon Talmor and Jonathan Berant. 2018.
\newblock \href {https://doi.org/10.18653/v1/N18-1059} {The web as a
  knowledge-base for answering complex questions}.
\newblock In \emph{Proceedings of the 2018 Conference of the North {A}merican
  Chapter of the Association for Computational Linguistics: Human Language
  Technologies, Volume 1 (Long Papers)}, pages 641--651, New Orleans,
  Louisiana. Association for Computational Linguistics.

\bibitem[{Wang et~al.(2020)Wang, Cho, and Lewis}]{wang-etal-2020-asking}
Alex Wang, Kyunghyun Cho, and Mike Lewis. 2020.
\newblock \href {https://doi.org/10.18653/v1/2020.acl-main.450} {Asking and
  answering questions to evaluate the factual consistency of summaries}.
\newblock In \emph{Proceedings of the 58th Annual Meeting of the Association
  for Computational Linguistics}, pages 5008--5020, Online. Association for
  Computational Linguistics.

\bibitem[{Weissenborn et~al.(2017)Weissenborn, Wiese, and
  Seiffe}]{weissenborn-etal-2017-making}
Dirk Weissenborn, Georg Wiese, and Laura Seiffe. 2017.
\newblock \href {https://doi.org/10.18653/v1/K17-1028} {Making neural {QA} as
  simple as possible but not simpler}.
\newblock In \emph{Proceedings of the 21st Conference on Computational Natural
  Language Learning ({C}o{NLL} 2017)}, pages 271--280, Vancouver, Canada.
  Association for Computational Linguistics.

\bibitem[{Welbl et~al.(2018)Welbl, Stenetorp, and
  Riedel}]{welbl-etal-2018-constructing}
Johannes Welbl, Pontus Stenetorp, and Sebastian Riedel. 2018.
\newblock \href {https://doi.org/10.1162/tacl_a_00021} {Constructing datasets
  for multi-hop reading comprehension across documents}.
\newblock \emph{Transactions of the Association for Computational Linguistics},
  6:287--302.

\bibitem[{Williams et~al.(2018)Williams, Nangia, and
  Bowman}]{williams-etal-2018-broad}
Adina Williams, Nikita Nangia, and Samuel Bowman. 2018.
\newblock \href {https://doi.org/10.18653/v1/N18-1101} {A broad-coverage
  challenge corpus for sentence understanding through inference}.
\newblock In \emph{Proceedings of the 2018 Conference of the North {A}merican
  Chapter of the Association for Computational Linguistics: Human Language
  Technologies, Volume 1 (Long Papers)}, pages 1112--1122, New Orleans,
  Louisiana. Association for Computational Linguistics.

\bibitem[{Wolf et~al.(2020)Wolf, Debut, Sanh, Chaumond, Delangue, Moi, Cistac,
  Rault, Louf, Funtowicz, Davison, Shleifer, von Platen, Ma, Jernite, Plu, Xu,
  Le~Scao, Gugger, Drame, Lhoest, and Rush}]{wolf-etal-2020-transformers}
Thomas Wolf, Lysandre Debut, Victor Sanh, Julien Chaumond, Clement Delangue,
  Anthony Moi, Pierric Cistac, Tim Rault, Remi Louf, Morgan Funtowicz, Joe
  Davison, Sam Shleifer, Patrick von Platen, Clara Ma, Yacine Jernite, Julien
  Plu, Canwen Xu, Teven Le~Scao, Sylvain Gugger, Mariama Drame, Quentin Lhoest,
  and Alexander Rush. 2020.
\newblock \href {https://doi.org/10.18653/v1/2020.emnlp-demos.6} {Transformers:
  State-of-the-art natural language processing}.
\newblock In \emph{Proceedings of the 2020 Conference on Empirical Methods in
  Natural Language Processing: System Demonstrations}, pages 38--45, Online.
  Association for Computational Linguistics.

\bibitem[{Yang et~al.(2018)Yang, Qi, Zhang, Bengio, Cohen, Salakhutdinov, and
  Manning}]{yang-etal-2018-hotpotqa}
Zhilin Yang, Peng Qi, Saizheng Zhang, Yoshua Bengio, William Cohen, Ruslan
  Salakhutdinov, and Christopher~D. Manning. 2018.
\newblock \href {https://doi.org/10.18653/v1/D18-1259} {{H}otpot{QA}: A dataset
  for diverse, explainable multi-hop question answering}.
\newblock In \emph{Proceedings of the 2018 Conference on Empirical Methods in
  Natural Language Processing}, pages 2369--2380, Brussels, Belgium.
  Association for Computational Linguistics.

\bibitem[{Zhu et~al.(2021)Zhu, Lei, Wang, Zheng, Poria, and
  Chua}]{zhu2021retrieving}
Fengbin Zhu, Wenqiang Lei, Chao Wang, Jianming Zheng, Soujanya Poria, and
  Tat-Seng Chua. 2021.
\newblock Retrieving and reading: A comprehensive survey on open-domain
  question answering.
\newblock \emph{arXiv preprint arXiv:2101.00774}.

\end{thebibliography}
\bibliographystyle{acl_natbib}

\clearpage

\appendix

\section{Details on Writing, Speed Validation, and Untimed Validation}
\label{app:ui}

\begin{figure*}[th!]
     \centering
         \includegraphics[width=0.95\textwidth]{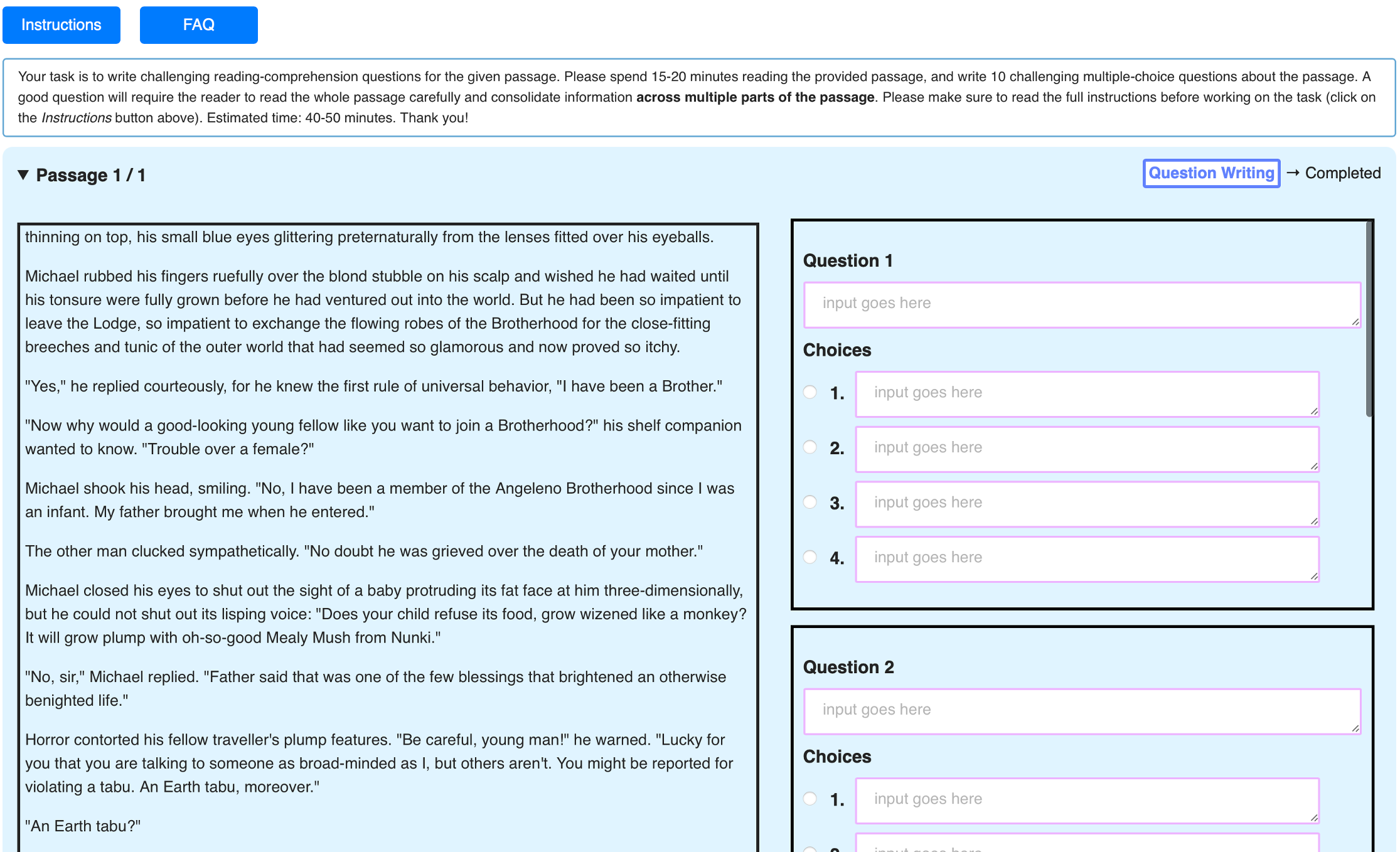}
    \caption{The writing UI.}
    \label{fig:ui_writing}
\end{figure*}

\subsection{Writing}

\subsubsection {Writer Recruitment}
\label{app:writer_recruitment}

We need writers who have good reading comprehension skills and/or writers who have experience constructing reading comprehension questions (e.g., literature teachers who have experience writing tests for their students). 
We hire two groups of writers on the freelancing platform Upwork (the second group two months after the first group, after we decided to increase the final size of the dataset). 

For each group, we advertise a task titled \textit{Writing college-level reading comprehension questions}. The job post is visible to all U.S. Upwork freelancers, and we specifically send out job invitations to promising freelancers who are writers or teachers, or who have college-level degrees in English, Literature, Creative Writing, Philosophy, Education, or similar fields. 
In the original job ad, we explain who we are and tell them how we will use their data. 
Specifically, we include the following phrase: ``The data we collect through this project will be made publicly available for AI research. We will not distribute any identifying information about you, the writers.''\footnote{We later obtained consent from writers who chose to allow us to name them in the Acknowledgments section.}

For the first group, we received 104 applications in the span of two weeks. Of those, we selected 26 people to complete a qualification task as a paid interview. For the second group, we received 65 applications and interviewed 11.
The interview task consists of (i) reading through detailed instructions, (ii) reading through a tutorial example passage with 10 example questions, each with an explanation of what made it a good or a bad question, and (iii) writing 10 reading comprehension questions for a new passage; regardless of whether we eventually hire them, we pay workers \$30 and estimate that this task takes 2 hours to complete. 
Three authors (of this paper) then assess each writer's work using the following criteria: (i) whether the writer-provided correct answers are actually the correct answers, (ii) whether the questions are answerable and unambiguous, and (iii) whether more than just a few sentences of context are needed to correctly answer the questions. 
Based on these criteria, we select the top performing 15 writers to continue on to the main task in the first group, and 7 in the second group. 

Of the 22 writers we hire after the interview, 15 have a college degree in English, literature, philosophy, creative writing, or education; 4 of these 15 writers are Ph.D. students or graduates. 11 of the 22 writers have taught high-school or college-level English or literature classes; among these 11 writers, 7 have 5+ years of teaching experience. 2 of the 22 writers mention that they write novels. 

\subsubsection{Writing Task}
\label{app:writing_task}

%Bonuses on top of the base rates for a single project are a feature of Upwork. Freelancers also value public project testimonials which are shown to future clients. We use these two media to incentivize workers to write high-quality questions. 

Each writer constructs 10 questions for a given passage, completing 6-30 passages in a given round and continuing for three complete rounds.\footnote{On average, group 1 writers complete 6, 14, 30 articles for the three batches, respectively; group 2 writers complete 6, 14, 20 articles for the three batches, respectively. The writing time limits for batches 1, 2, 3 are around 1, 2, 3 weeks, respectively. Validation for any batch takes less than a week.} 
Each round is followed by feedback (detailed below) to allow writers to improve for the next round.
Writers earn \$12.50 per passage and receive a bonus of \$1.20 for each question that meets the following criteria: (i) the majority of validators agree with the writer's original label, (ii) the majority of validators rated the question as answerable and unambiguous, and (iii) the majority of validators answered the question \textit{incorrectly} in the speed validation task (\S\ref{sec:speed-val}).
On average, writers receive bonuses on 4.2 questions per passage, resulting in average earnings of \$17.54 per passage. Based on writer self-reports, the median time to complete one writing task is about 50 minutes, for an effective rate of \$21.05/hr. {Upwork charges fees on the workers' end. We account for this by adding an extra 20\% to their pay, bringing our final cost to \$2.10 per question.} 

Besides using the monetary bonus as an incentive for writing answerable, unambiguous, and difficult questions, we also instruct writers that their questions should use the entire context. Throughout the course of data collection, we provide writers with detailed feedback based on validations (detailed in \S \ref{sec:untimed_val}) and this feedback includes information about how much of the passage needed to be read in order to answer the question. We monitor the proportion of questions that require more than a few paragraphs of context to answer correctly; if this rate significantly lags behind other writers, we inform the writers that their work is falling below expectations and ask them to be more careful with this issue in the next round. 
We also encourage writers to write difficult distractors, and the feedback we provide also contains what annotators think is the most difficult distractor for each question (\S \ref{sec:untimed_val}). 

If writers have fewer than 40\% of questions meet the above three bonus criteria \emph{and} fewer than 75\% of questions meet criteria (i) and (ii), we exclude them from future writing rounds: One writer was excluded after batch 1, and one writer was excluded after batch 2, for this reason. We also exclude two writers who missed deadlines by significant margins. Two other writers voluntarily left the project before finishing all three batches.

\subsubsection{The Writing UI}
\label{app:writing_ui}
Figure~\ref{fig:ui_writing} shows our writing UI. A writer creates 10 multiple-choice questions with four answer options each on each page. Before the interview task and each batch of data collection, we explain our bonus structure to the writers. In order to encourage writers towards writing the types of questions that require understanding of the general context from the passage, we provide the following examples of themes that questions can target in order to spur writers' creativity and provide suggestions if they have trouble coming up with difficult questions; however, they do not have to follow our suggestions. 
\begin{itemize}[noitemsep]
    \item Characters' feelings and motivations
    \item Causes and consequences of described events
    \item Definitions, properties, and processes explained in a passage
    \item The summary and lesson of a passage
    \item What would have happened had a character made a different choice
\end{itemize}

We also allow writers to skip a given passage in case they find that they would be unable to write high-quality questions for that passage. Specifically, we tell writers the following. 
\begin{quote}
If a passage is too difficult to write questions for, you can skip the article by choosing another URL to work on. We recommend that you do this if: (1) The text is hard to read due to major formatting issues. (2) The text is very technical or relies on cultural knowledge that you’re unfamiliar with. (3) You think the passage is much too boring. We ideally want you to write questions for passages you find interesting!
\end{quote}

\begin{figure*}[th!]
     \centering
         \includegraphics[width=0.95\textwidth]{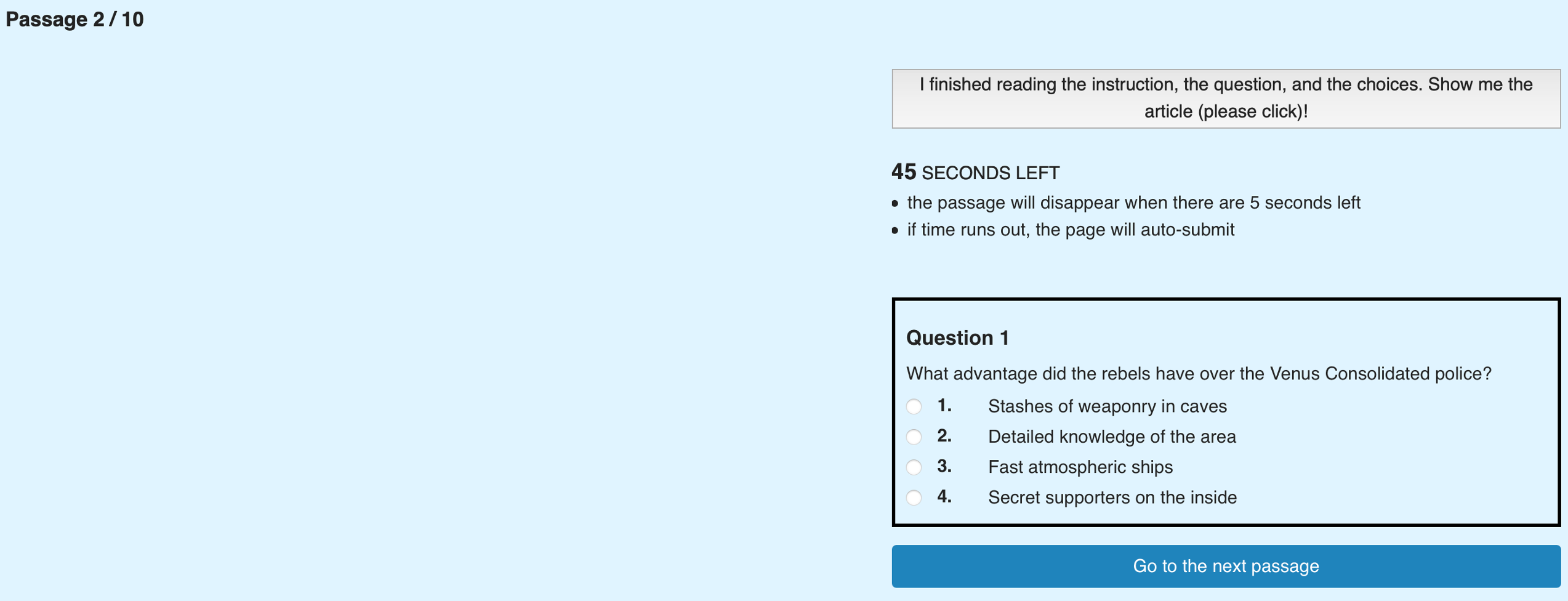}
    \caption{The speed validation UI before clicking the top-right button.}
    \label{fig:ui_speed_val_1}
\end{figure*}

\begin{figure*}[th!]
     \centering
         \includegraphics[width=0.95\textwidth]{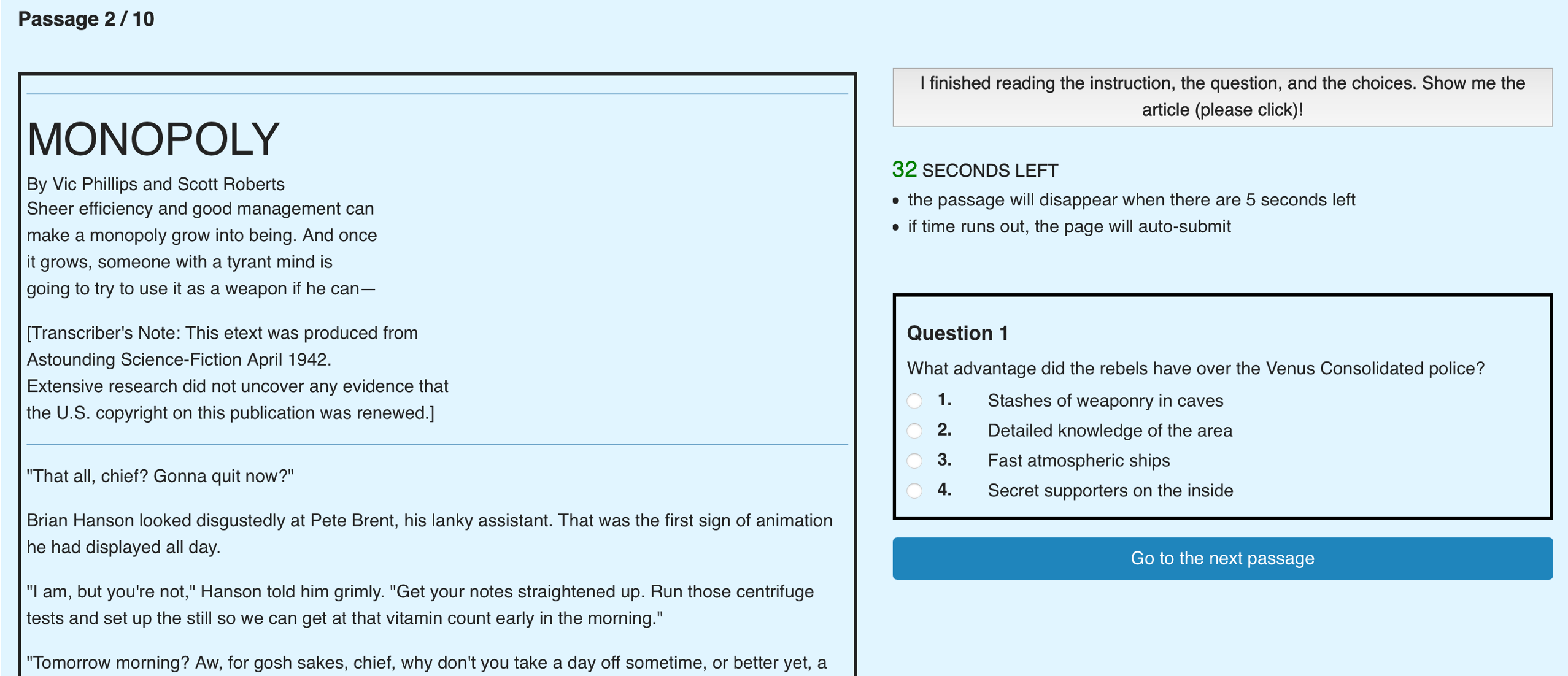}
    \caption{The speed validation UI after clicking the top-right button.}
    \label{fig:ui_speed_val_2}
\end{figure*}

\subsection{Validation}

\subsubsection{Annotator Recruitment}
\label{app:annotator-recruitment}

% qualification task
We recruit annotators via Amazon Mechanical Turk (MTurk).
We use a qualification task to identify annotators with good reading comprehension skills.
This task is open to all workers with more than 1000 tasks (HITs) accepted and a HIT accept rate of at least 98\%.
We pay \$5 for completing the qualification task, plus a \$5 bonus for passing it.
The task consists of a $\sim$3000-word passage with 10 multiple choice questions written or reviewed by the authors, each with a series of evaluation questions asking about the quality of that question.
Of the 10 questions, 2 are intentionally ambiguous\footnote{We later found that one question was \textit{unintentionally} ambiguous; we do not use this question in assessing whether workers pass the qualification.} in order to test if workers can accurately identify poorer quality questions.

% criteria to get the qualification
In order to pass the qualification, workers need to (i) get at least 6/7 or 7/8 of the unambiguous questions correct, (ii) correctly identify at least one of the two ambiguous questions as ambiguous/unanswerable, and (iii) correctly identify at least half of the unambiguous questions as unambiguous.
A total of 148 crowdworkers completed the task, and 45 of them passed (30.4\%).
All workers who pass the qualification are invited to complete tasks as part of both the speed validation and the untimed validation. 
We make it clear in this task as well as the main speed validation and untimed validation tasks who we are and which research group we are affiliated with. 
In order to help workers understand that we plan to use their data for research purposes related to language technologies, we also include the following in the FAQ section of each hit: ``With your help, we think we'll be able to build some pretty exciting technologies to help computers better understand human language.'' 

\subsubsection{Speed Validation}
\label{app:speed_val}

\paragraph{Catch Questions}
We expect accuracy in the speeded task to be fairly low, so we construct catch questions to ensure that workers are not randomly guessing without attempting to find the correct answer.
These trials are written by the authors and are designed to be answerable with only 45 seconds of access to the passage. 
For example, a catch question may ask who spoke a quote, like ``Who said `You're a wizard Harry!'?'', where a single ctrl+F search of the quote gives the annotator the answer. Another catch question may have four options, three of which are clearly improbable. 
We do not validate the catch questions for correctness in the untimed validation, and so we do not include them in the final dataset, but we release them as a supplemental file for reproducibility.

\paragraph{Payment}
For most of the tasks, we pay workers \$2.25 per HIT and award a bonus of \$0.20 for each correct answer. 
However, during the first of six rounds, we paid \$2.00 per HIT with a \$0.18 bonus for each correct answer. 
After asking workers for feedback about the task via a survey, we decided to increase the rate of pay because workers reported spending slightly longer on the task than we originally estimated.

\paragraph{Task Procedure}
Each MTurk task consists of 10 speed validation questions from different randomly chosen articles. In each task, once the annotator clicks into the page, they have unlimited time to read the question and the answer options, but the article is not shown (Figure~\ref{fig:ui_speed_val_1}).
Then, the annotator clicks the button that says ``I finished reading the instruction, the question, and the choices. Show me the article (please click)!'' As soon as the annotator clicks the button, the countdown clock of 45 seconds starts, and the article appears (Figure~\ref{fig:ui_speed_val_2}). The annotator can make the choice and submit at any time. 

When there are only 5 seconds left, the article hides itself. The annotator has 5 seconds to make the choice. If the time expires, the page auto-submits, and we record that the annotator did not make a choice, which we score as incorrect. The exact instructions that the annotator sees are as follows: 
\begin{quote}
    In this task, you will see a long text passage and a multiple choice question that can be answered from that text. Read the question and select the best answer option. You only have \textbf{45 seconds} to choose an answer, so this is not enough time to read the whole passage. We encourage you to skim and use keyword-based searches (e.g., using ctrl+F) during this time. Even if you are unsure of the answer, you should make an educated guess.

    After 45 seconds, your answer will be locked in and submitted. If you have not provided an answer at the end of 45 seconds, you will not be able to answer this question and will be automatically moved to the next question. We will not reject your work for a couple of blank answers, but excessive failures to answer will result in a loss of the qualification to complete these HITs.

    You will not be penalized for wrong answers. We will give you a bonus of \textbf{\$0.20} for each correct answer. Thus, it is in your best interest to attempt each question, even if it is just a guess. A few of the questions will be answerable in just 45 seconds, and we do expect you to get these right reasonably often.
\end{quote}

\paragraph{Annotator Performance}
Individual annotators consistently scored well above the chance rate of 25\% on the catch questions.
In all cases where an annotator's accuracy fell below 50\% in a round, they were removed from future rounds. Two annotators fell below this threshold, though in that round they had also performed below threshold in the untimed validation. No annotators needed to be removed \textit{solely} based on performance in the speed validation task.
Average overall accuracy on the catch questions was 83.8\%, indicating that most workers were able to develop a strategy for finding a correct answer when it could be found.

Accuracy on the questions written by Upwork writers was 48.2\% overall, but annotators got better at this task over time, likely by developing new strategies to search for answers.
Average accuracy was 39.5\% in the first round, rising to 58.4\% in the final round of data collection.
When the majority of annotators (at least 3/5) are able to answer questions correctly in this setting, we exclude that question from the \textsc{hard} subset.

\begin{figure*}[th!]
     \centering
         \includegraphics[width=0.95\textwidth]{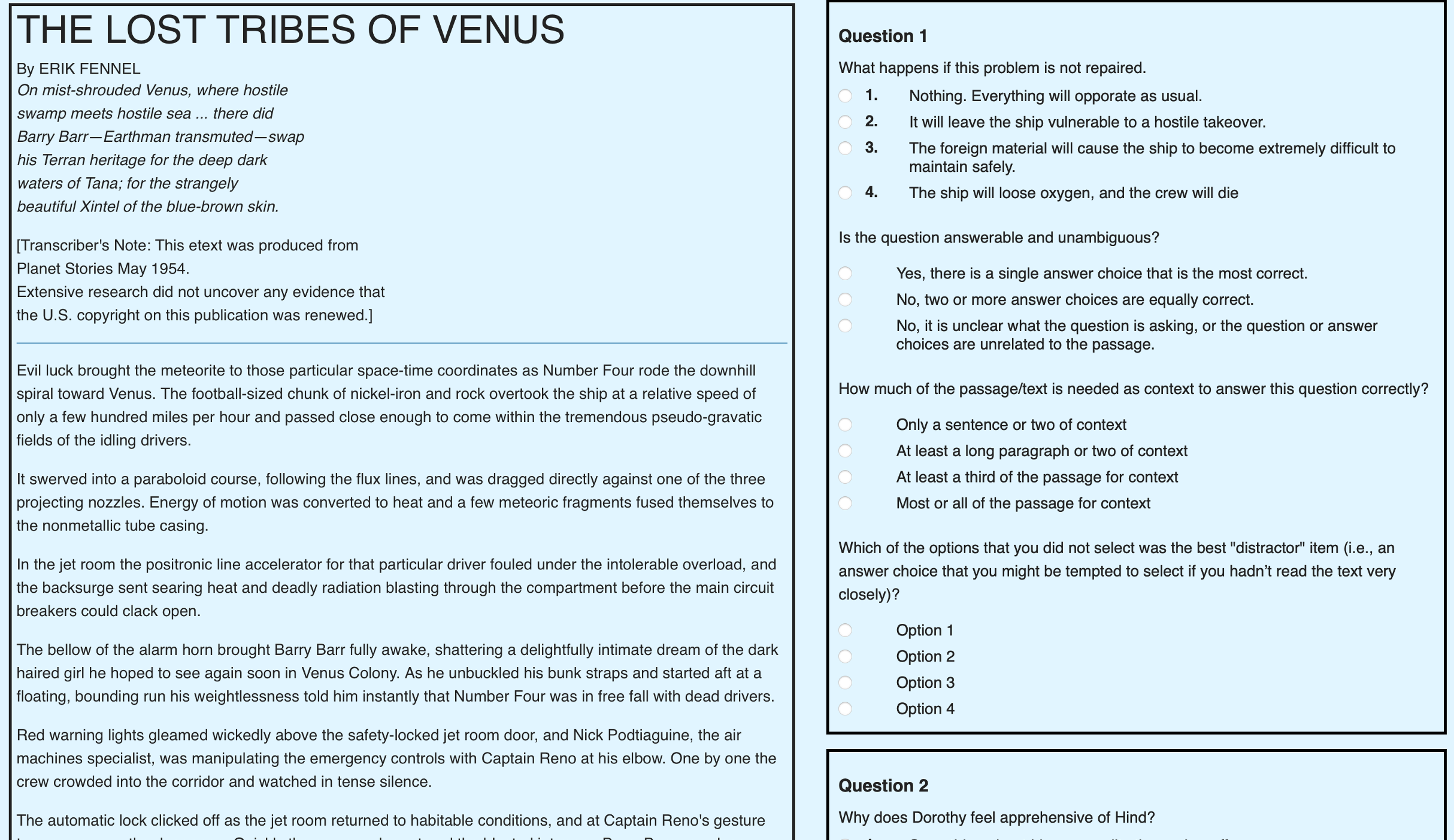}
    \caption{The (untimed) validation UI.}
    \label{fig:ui_val}
\end{figure*}

\subsubsection{Untimed Validation}
\label{app:untimed_val}

Figure~\ref{fig:ui_val} shows the UI for untimed validation. As two writers each write 10 questions for the same article, there are 20 unique questions per article. Each validation UI page contains all 20 sets of questions, and each set of questions contains the reading comprehension question and the three additional evaluation prompts. Therefore, in total, there are 80 prompts on each page. The annotator has to complete all 80 before they can submit the page and complete the task. The exact instructions that the annotators read for this task are as follows:
\begin{quote}
    In this task, you will answer multiple choice questions corresponding to a long article. Each passage comes with 20 sets of questions. Each set contains a comprehension question and three evaluation questions. Please read each passage carefully before answering the questions for that passage. Estimated time: 45-55 minutes.

	If it is impossible to say which of the other answer options is correct, then select the answer option that is \textit{closest} to correct.

	You will receive a bonus of \$0.50 for each reading comprehension question that you correctly answer. We consider the answer to be correct if your response of the reading comprehension question and the first evaluation question \textbf{both} agree with the most common answer from other workers and the original writer. If there's no agreement on an answer, we count it as correct for everyone. This means that it's possible to receive a total bonus of \$10.00 on this HIT. 

	We expect you to answer most of the questions correctly; however, we understand that some questions may be difficult, ambiguous, or mal-formed. If you answer a large number of questions incorrectly, we will disqualify you from future work on this task.  

	For the second and third evaluation questions, we will check if your choices agree with the majority of other workers who work on the same task, but your bonus is not dependent on these answers. The results will also be used to determine whether you retain the qualification for future batches of this task, but to a lower threshold because these questions tend to be more subjective.
\end{quote}

\begin{table*}[th]
\setlength{\tabcolsep}{3.0pt}
\centering
\small
\begin{tabular}{lrrrl}
    \toprule
    Question Type & \# \textsc{easy} & \# \textsc{hard} & \% total & Example from the Training Set \\
    \midrule
    what & 1361 & 1471 & 42.2 & What is the immediate significance of Ed defending the ads on his Facebook? \\
    why & 832 & 825 & 24.6 & Why does Howell not want Linton to approach Snead in the restaurant? \\
    how & 385 & 416 & 11.9 & How does Barker view his own film? \\
    which & 253 & 244 & 7.4 & Which word least describes McGill? \\
    who & 151 & 132 & 4.2 & Who is the most hated celebrity of 1999? \\
    how + meas. & 51 & 75 & 1.9 & How many caves had Garmon and Rolf traveled through before their crash? \\
    yes/no & 53 & 55 & 1.6 & Was it Nelson's decision to become part of the military? \\
    where & 43 & 42 & 1.3 & Where was the space craft heading in the end? \\  % Where is the Farm? \\
    when & 35 & 34 & 1.0 & When did the Hanseatic League begin? \\
    other & 155 & 124 & 4.1 & \makecell[tl]{Dole's quote would have been perceived as \_\_\_\_\_\_ if it had included included \\ the exclamation points from his tone?} \\ % The changes that Ninian make in Martin's life \\
    \bottomrule
\end{tabular}
\caption{Different question types in QuALITY, split by \textsc{hard} and \textsc{easy} subsets. 
`How + meas.' collapses multiple questions with `how' plus some measurement, such as `how long' or `how many.' 
}
\label{tab:questiontypes}
\end{table*}

\paragraph{Annotator Performance}
% average performance
Individual annotator agreement with the gold label is 91.2\% for all data collected in the main data collection (not including the responses collected to measure human accuracy described in \S \ref{sec:human-acc}).
% criteria to keep the qualification
Throughout the course of the study, workers need to maintain at least 75\% accuracy each round to keep the qualification and continue to the next round.
In a few cases, we identify passages that are themselves ambiguous or especially difficult. In these cases, we do not use those passages in computing by-round accuracy for the annotators.
% number of annotators at each round
We exclude a total of 11 workers throughout the course of data collection for low accuracy, most of them after the first or second round.

\paragraph{Data Reannotation}
% automatic reannotation
During each untimed validation round, we keep track of the rate at which each worker agrees with the original writers' labels for each question in order to quickly identify cases where either (i) a worker has misunderstood the passage, or (ii) a worker is putting insufficient effort towards the task.
For any tasks where the individual annotator disagrees with the writer's labels on at least 40\% of questions, we automatically re-post that passage for reannotation and replace the data, with the assumption that the annotator may have misunderstood something crucial in the passage.\footnote{We identified 10 passages that, after multiple rounds of reannotation, were not passing this threshold. This may be due to one of the writers misunderstanding the passage and thereby creating several ambiguous questions, so for these 10 passages, we chose to include \textit{all} annotations collected rather than replace annotators' data.}  
% maj. vote based reannoation
After all the annotations are complete, we calculate the gold label answer via majority vote of annotators plus the original writer's label, and assess individual annotator accuracy. 
If any worker is excluded in a round for low accuracy (i.e., below 75\% accuracy), we discard all of their responses from that round, reannotate and replace their data, and re-calculate the gold label and accuracy scores.

\section{Data}
\label{app:data}

\paragraph{Switchboard Data}
In order to increase the diversity of genres we use as context passages, we attempted to include Switchboard conversations. However, after presenting just 12 such conversations to our writers, we decided to discard all the Switchboard questions because many writers informed us that it was very difficult to come up with difficult questions for the Switchboard conversations.
The writers indicated they found the Switchboard articles more difficult because the conversations are relatively short and usually involve very simple everyday topics, without the kinds of plot twists that are more common in short stories or complex details that are more common in long-form articles. 

\begin{figure}[t!]
        \centering
        \includegraphics[scale=0.49]{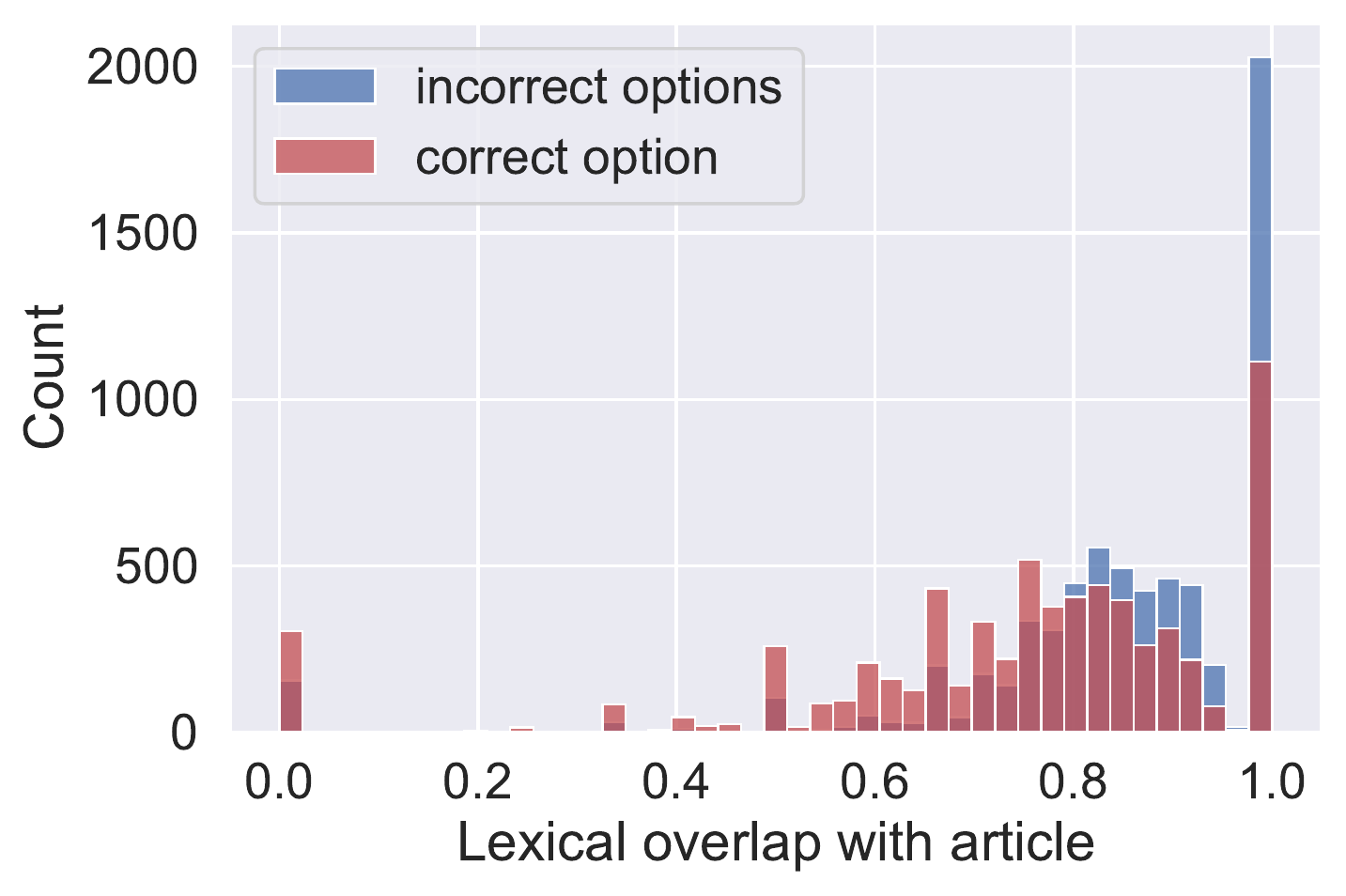}
    \caption{Lexical overlap of the correct and incorrect options with the context. Since each question has three incorrect options, we use the option with the highest lexical overlap.} 
    \label{fig:lexical_overlap_max}
\end{figure}

\paragraph{Lexical Overlap}
We analyze the lexical overlap between the answer options and the passage text (detailed in \S \ref{ssec:overlap}). 
In Figure \ref{fig:lexical_overlap_max}, we plot the lexical overlap of both the correct answer option and the incorrect answer option with the passage. 

%\section{Question analysis}
\subsection{Question Types}
\label{app:questiontypes}

As described in \S \ref{sec:question-types}, we analyze the different question types in QuALITY, split by \textsc{hard} and \textsc{easy} subsets, and present these results along with examples in Table \ref{tab:questiontypes}. 
Most of the questions in the `other' category are finish-the-phrase style questions, and for the example in the table, the answer options are different ways that sentence could be completed. 
Note that in most of the yes/no questions, the answer options include the necessary reasoning to support the yes/no answer, meaning that these are often complex, multi-part questions. 
Examples shown in Table \ref{tab:questiontypes} are randomly selected from the training set, with the caveat that we selected the `when' question by hand, since about half of the questions categorized as `when' are referencing some timepoint (e.g., `when X happened, what did ...')

\begin{table*}[th]
\setlength{\tabcolsep}{3.0pt}
\centering
\small
\begin{tabular}{p{30ex}p{63ex}p{15ex}}
    \toprule
     Question & Answer Options & Reasoning Types \\
    \midrule
     What would have happened if Click’s camera broke in the crash? & (a) Irish would have died on impact. (b) They would have returned immediately to Luna Base. (c) They would have caught Gunther faster. (d) They would have continued to believe the monsters were real. & What if; Event \\
    \midrule
     What isn't a reason for narrator to be so skeptical of Gorb? & (a) Gorb looked just like an Earthling (b) Gorb was asking for too much money (c) Gorb had no proof to back up his claims (d) he had never heard of Wazzenazz & Why/reason; not/except \\
    \midrule
     How many caves had Garmon and Rolf traveled through before their crash? & (a) thirty seven (b) forty seven (c) thirty (d) forty & Numeric \\
    \midrule
     How do you think Meredith feels about the rest of the crew? & (a) She has a close bond of respect and (platonic) love for the rest of the members (b) She respects and loves one person the most (c) She's become friends with them slowly over time and appreciates them all (d) She respects one person the most and loves another person the most & Description; symbolism/terpretation; relation \\
    \midrule
     The less you share... & (a) ...the more privacy you have. (b) ...the more your intellectual property is protected. (c) ...the less power you have. (d) ...the less your cultural goods will be appropriated. & Symbolism/in-terpretation; finish the phrase \\
    \midrule
    How did Meryl Streep prepare for the role of Roberta? & (a) She learned to play the violin without any former instrument training. (b) She began to act very helplessly and feeble around the rest of the cast. (c) She is a method actor and became very vulnerable. (d) She made herself look dumpy and thick-waisted. & How/method \\
    \bottomrule
\end{tabular}
\caption{Full examples of the annotations from our analysis of reasoning types on a subset of questions from QuALITY. Examples are taken from analyzed examples from the training set. Examples are selected non-randomly and are intended to demonstrate a range of reasoning types observed.}
\label{tab:reasoning_examples}
\end{table*}

\subsection{Reasoning Types}
\label{app:reasoningtypes}

For the purposes of this analysis, we define reasoning type as the category that needs to be reasoned about in selecting the correct answer option (e.g., `person' is usually a `who' question and corresponds to answer options that are characters or people) or the type of strategy that must be used in answering (e.g., `symbolism/interpretation' requires the reader to extrapolate from the context or identify something not stated in a passage, like its theme).
We identify 15 categories of reasoning types to include in our analysis.
These categories are initially inspired by those used in NarrativeQA, but we adapt them to our dataset, as we find that many questions in QuALITY do not fit their categorization.
These categories are not mutually exclusive, and nearly a third of the questions are categorized as two or more types.

\begin{table*}[th]
\setlength{\tabcolsep}{3.6pt}
\centering
\small
\begin{tabular}{llcccccc}
    \toprule
    & & & \multicolumn{3}{c}{Extraction Based on Qs}  \\
    \cline{4-6} 
    \noalign{\smallskip}
    Training Data & Model & Full & R-1 & fastText & DPR & {Question-Only} \\ 
    \midrule
    QuALITY 
      & Longformer-base & 30.7\ /\ 29.3 & -- & -- & -- & -- \\
      % & LED-base & 25.3\ /\ 25.8 & -- & -- & -- & -- \\
      & LED-base & 25.1\ /\ 24.6 & -- & -- & -- & -- \\
      & LED-large & 24.2\ /\ 24.5 & -- & -- & -- & -- \\
      & RoBERTa-base & -- & 33.4\ /\ 30.7 & 39.7\ /\ 36.1 & 39.9\ /\ 34.0 & 36.6\ /\ 34.8 \\
      & RoBERTa-large & -- & 29.4\ /\ 28.0 & 42.7\ /\ 35.7 & 26.2\ /\ 25.1 & 26.4\ /\ 25.7 \\
      & DeBERTaV3-base & -- & 36.7\ /\ 35.7 & 38.9\ /\ 35.9 & 44.1\ /\ 38.5 & 38.2\ /\ 35.6 \\
      & DeBERTaV3-large & -- & 46.5\ /\ 39.3 & 45.5\ /\ 40.2 & 49.0\ /\ 41.2 & 39.7\ /\ 35.2 \\
      & T5-base & -- & 28.0\ /\ 28.0 & 28.9\ /\ 27.4 & 29.3\ /\ 29.1 & 30.1\ /\ 29.9 \\
    \midrule 
    RACE 
      & Longformer-base & 35.2\ /\ 30.8 & -- & -- & -- & -- \\
      & RoBERTa-base & -- & 42.4\ /\ 36.8 & 43.2\ /\ 37.2 & 44.2\ /\ 36.1 & 33.8\ /\ 29.7  \\
      & RoBERTa-large & -- & 47.0\ /\ 37.5 & 47.9\ /\ 40.2 & 48.7\ /\ 40.2 & 36.6\ /\ 33.1   \\
      & DeBERTaV3-base & -- & 45.3\ /\ 36.1 & 46.1\ /\ 39.0 & 47.8\ /\ 39.4 & 34.7\ /\ 30.5 \\
      & DeBERTaV3-large & -- & 52.9\ /\ 43.4 & 51.2\ /\ 42.4 & 53.0\ /\ 44.4 & 36.5\ /\ 30.0 \\
      & T5-base & -- & 41.5\ /\ 38.6 & 42.3\ /\ 39.9 & 43.4\ /\ 41.0 & 36.5\ /\ 34.8 \\
    \midrule 
    \multirow{3}{2cm}{RACE \\ \ \ \ \ $\downarrow$ \\ QuALITY}
      & Longformer-base & 39.5\ /\ 35.3 & -- & -- & -- & -- \\
      & LED-base & 37.2\ /\ 33.8 & -- & -- & -- & -- \\
      & LED-large & 39.4\ /\ 35.3 & -- & -- & -- & -- \\
      & RoBERTa-base & -- & 42.1\ /\ 38.3 & 43.0\ /\ 40.1 & 44.3\ /\ 39.8 & 38.1\ /\ 37.5 \\
      & RoBERTa-large & -- & 48.0\ /\ 40.8 & 50.4\ /\ 43.7 & 51.4\ /\ 44.7 & 40.4\ /\ 37.1 \\
      & DeBERTaV3-base & -- & 46.8\ /\ 38.7 & 49.8\ /\ 43.2 & 51.2\ /\ 42.4 & 41.4\ /\ 37.9  \\
      & DeBERTaV3-large & -- & 53.8\ /\ 46.3 & 54.7\ /\ \bf{46.7} & {\bf{55.4}}\ /\ 46.1 & 43.3\ /\ 38.2 \\
     & T5-base & -- & 41.1\ /\ 40.1 & 40.8\ /\ 40.1 & 41.6\ /\ 39.8 & 36.4\ /\ 35.9 \\
      \midrule
      -- & Human Annotators & 93.5\ /\ 89.1 & -- & -- & -- & -- \\
    \bottomrule
\end{tabular}
\caption{Accuracy on the full QuALITY test set and the QuALITY-\textsc{hard} subset (formatted as full\ /\ \textsc{hard}). The ``Full'' column has results from training with the source inputs truncated to fit into memory. %, without using an extractive model to select portions of text. 
R-1 (ROUGE-1), fastText, DPR are three extraction methods (\S \ref{sec:models}) used to select relevant portions of the source text. 
% Poor RoBERTa-large performance (for training on QuALITY only) is likely due to unstable training given our small training set \citep{mosbach2021on}.
Results for training on QuALITY and RACE$\rightarrow$QuALITYare identical to Table \ref{tab:main-results-noRACE} in the main text, this table simply presents those results alongside results from training only on RACE.
}
\label{tab:main-results}
\end{table*}

\paragraph{Reasoning Type Definitions}

The following includes definitions of all the categories used, along with at least one hand-selected example to demonstrate a question belonging to that category.
All in-text examples are selected out of the training set.

\begin{itemize}
    \item \textbf{Description}: The question relies on the reader reasoning about which description is correct. Often these questions are about describing a character's feelings (`How do Lowry and the Exec feel about the Venusians?') or point of view (`How is the book "Living a Normal Sex Life" seen by these people?'), describing a feature of the story (`What makes Grannie Annie's writing remarkable?'), or describing an individual (`Which word least describes Don?')
    \item \textbf{Why/reason}: The question relies on the reader reasoning about the cause or explanation for something in the story. Most of these questions begin with `why' and ask about the cause of an event (`Why does the crew get off the ship with Moran?'), causes of characters' feelings (`Why does Ben take offence to Cobb's comments about spacemen?'), or characters' internal motivations (`Why does Joseph lie about the water supply?'), though other questions formulate this differently while still asking for the underlying reason (`What makes Gubelin an outlier in the present day?' and `What is the purpose of a comanalysis?').
    \item \textbf{Symbolism/interpretation}: The question relies on the reader making an interpretation that goes beyond what is explicitly said in the story, or it asks about symbolism or themes from the story. Many questions explicitly ask the reader to interpret what message the author was trying to convey (`What point is being made by comparing Fight Club to the UFC?') or what tone the story takes (`What is the tone like throughout the story?'). Other questions require the reader to predict what will happen next (`What will happen next to Jery?') or ask about the use of literary cues like irony (`What is ironic about Earth's customer service policy?'). 
    \item \textbf{How/method}: The question relies on reasoning about how something happened or the method that was used. Most of these questions rely on the question word `how' to ask about a process (`How did Meryl Streep prepare for the role of Roberta?'), the manner in which something happens (`How did Templin find about about Pendleton's death?'), or a method by which something happens (`How does the shape of Starre's ship benefit them?').
    \item \textbf{Event}: The question relies on reasoning about an event, or asks for an event as the answer option. The majority of these questions focus on what someone did/plans to do (`What did Joe and Glmpauszn plan to do?') or what happened to someone (`What happened to Morgan Brockman by the end of the passage?').
    \item \textbf{Person}: The question relies on reasoning about which person or people are involved. Most ask about a specific person (`Who is Owen Fiss and what did he do?'), though many questions of this type still require reasoning about the entire passage to answer (`Who seems to have the least to hide in the text?').
    \item \textbf{Not/except}: The question requires the reader to select the answer option that \textit{least} answers the question, flipping the typical way a multiple-choice task is performed. All of these questions use some word to indicate this flipping, such as `least' (`Which word least describes McGill?'), `not' (`What word doesn't describe the natives from Tunpesh?'), or `except' (`Dole makes all of the following charges against the New York Times EXCEPT for:').
    \item \textbf{Relation}: The question relies on reasoning about the relationship between two or more characters, as in `Who is Sporr and what is his authority in calling the narrator Yandro?' or questions that ask about how one character feels about another (`How does Jakdane feel about Trella?').
    \item \textbf{Entity}: The question relies on reasoning about a non-human entity or a group, as in `We can assume that Saladin's army represents which group?'.
    \item \textbf{Finish the phrase}: The form of the question requires either a fill-in-the-blank style response or is a partial phrase that must be completed by selecting the correct answer option. Often, these questions do not include an explicit question word. Some of them have a blank written in (`The film reviewer is generally \_\_\_\_\_ the actors in "Princess Mononoke," and \_\_\_\_\_\_ the actors in "The Limey," respectively:') and others are just a partial sentence (`The less you share...').
    \item \textbf{Location}: The question relies on reasoning about a place, as in `What city is Temple-Tracy in?'.
    \item \textbf{Numeric}: The question relies on finding or computing the correct numeric option, as in `How many caves had Garmon and Rolf traveled through before their crash?'. 
    \item \textbf{Object}: The question relies on reasoning about an object, as in `What does Captain Hannah use as an organic processor?'.
    \item \textbf{What if}: The question requires the reader to make an inference about what \textit{would have been true} if some fact from the story were changed, and most of these questions explicitly set up the counterfactual scenario (`What would have happened if the Peace State had not crash landed?').
    \item \textbf{Duration}: The question relies on reasoning about how long something happened for or how much time passed between two events, as in `How long did Maggie care for Ben before he finally awoke after rescuing him?'.
\end{itemize}

\begin{table*}[th]
\setlength{\tabcolsep}{3.6pt}
\centering
\small
\begin{tabular}{llcccccc}
    \toprule
    & & & \multicolumn{3}{c}{Extraction Based on Qs}  \\
    \cline{4-6} 
    \noalign{\smallskip}
    Training Data & Model & Full & R-1 & fastText & DPR & {Question-Only} \\ 
    \midrule
    QuALITY 
      & Longformer-base & 33.7\ /\ 32.6 & -- & -- & -- & -- \\
      & LED-base & 25.1\ /\ 24.3 & -- & -- & -- & -- \\
      & LED-large & 25.1\ /\ 25.6 & -- & -- & -- & -- \\
      & RoBERTa-base & -- & 33.7\ /\ 32.0 & 39.2\ /\ 37.2 & 40.0\ /\ 36.4 & 36.0\ /\ 35.6  \\
      & RoBERTa-large & -- & 30.0\ /\ 28.7 & 42.7\ /\ 38.3 & 26.7\ /\ 24.0 & 26.7\ /\ 23.0 \\
      & DeBERTaV3-base & -- & 34.7\ /\ 33.0 & 38.0\ /\ 35.3 & 41.8\ /\ 37.4 & 36.9\ /\ 34.3 \\
      & DeBERTaV3-large & -- & 44.0\ /\ 37.6 & 44.3\ /\ 37.9 & 45.1\ /\ 39.2 & 38.1\ /\ 33.7 \\
      & T5-base & -- & 27.3\ /\ 26.6 & 27.6\ /\ 25.5 & 28.3\ /\ 29.4 & 27.9\ /\ 28.7 \\
    \midrule 
    RACE 
      & Longformer-base & 34.5\ /\ 31.6 & -- & -- & -- & -- \\
      & RoBERTa-base & -- & 43.7\ /\ 38.2 & 43.3\ /\ 38.9 & 44.1\ /\ 37.3 & 36.8\ /\ 34.6  \\
      & RoBERTa-large & -- & 48.6\ /\ 41.7 & 48.4\ /\ 42.3 & 50.9\ /\ 45.3 & 37.2\ /\ 35.3   \\
      & DeBERTaV3-base & -- & 46.5\ /\ 38.7 & 44.8\ /\ 37.9 & 48.8\ /\ 41.7 & 35.8\ /\ 31.6 \\
      & DeBERTaV3-large & -- & 51.2\ /\ 43.9 & 50.5\ /\ 43.8 & 53.5\ /\ 47.3 & 38.3\ /\ 34.3 \\
      & T5-base & -- & 39.0\ /\ 37.7 & 39.7\ /\ 39.2 & 39.9\ /\ 38.5 & 37.2\ /\ 35.6 \\
    \midrule 
    \multirow{3}{2cm}{RACE \\ $\downarrow$ \\ QuALITY}
      & Longformer-base & 38.1\ /\ 32.8 & -- & -- & -- & -- \\
      & LED-base & 35.6\ /\ 32.0 & -- & -- & -- & -- \\
      & LED-large & 39.9\ /\ 39.6 & -- & -- & -- & -- \\
      & RoBERTa-base & -- & 43.7\ /\ 38.2 & 41.7\ /\ 36.2 & 43.8\ /\ 37.2 & 37.4\ /\ 36.6 \\
      & RoBERTa-large & -- & 47.7\ /\ 42.5 & 46.8\ /\ 43.1 & 50.8\ /\ 46.2 & 39.1\ /\ 37.8 \\
      & DeBERTaV3-base & -- & 45.5\ /\ 40.0 & 46.6\ /\ 40.1 & 46.7\ /\ 40.9 & 39.6\ /\ 35.2 \\
      & DeBERTaV3-large & -- & 51.7\ /\ 44.7 & 50.7\ /\ 43.3 & 53.6\ /\ 47.4 & 41.4\ /\ 39.2 \\
      & T5-base & -- & 40.0\ /\ 38.2 & 40.4\ /\ 38.6 & 39.0\ /\ 37.9 & 37.1\ /\ 36.1 \\
    \bottomrule
\end{tabular}
\caption{Accuracy on QuALITY development set (full\ /\ \textsc{hard}). 
}
\label{tab:dev-results}
\end{table*}

\paragraph{Annotation Details}
Three authors of this paper analyze a set of 500 randomly selected questions.
One author annotates all 500, and the other two annotators analyze 250 each, such that each example is annotated by two unique individuals.
Following annotation, the authors discuss any disagreements and adjust their original coding once consensus is reached.
Using this consensus approach allows for clarification of the categories during and after annotation, which leads to an internally consistent coding scheme.

\paragraph{Sample Annotations}

Table \ref{tab:reasoning_examples} shows a set of representative example annotations from this analysis, demonstrating several sentences that were categorized as more than one reasoning type.

\section{More Details on Analysis}
\label{app:analysis}

\paragraph{Lexical Overlap}

\begin{figure}[th!]
        \centering
        \includegraphics[scale=0.5]{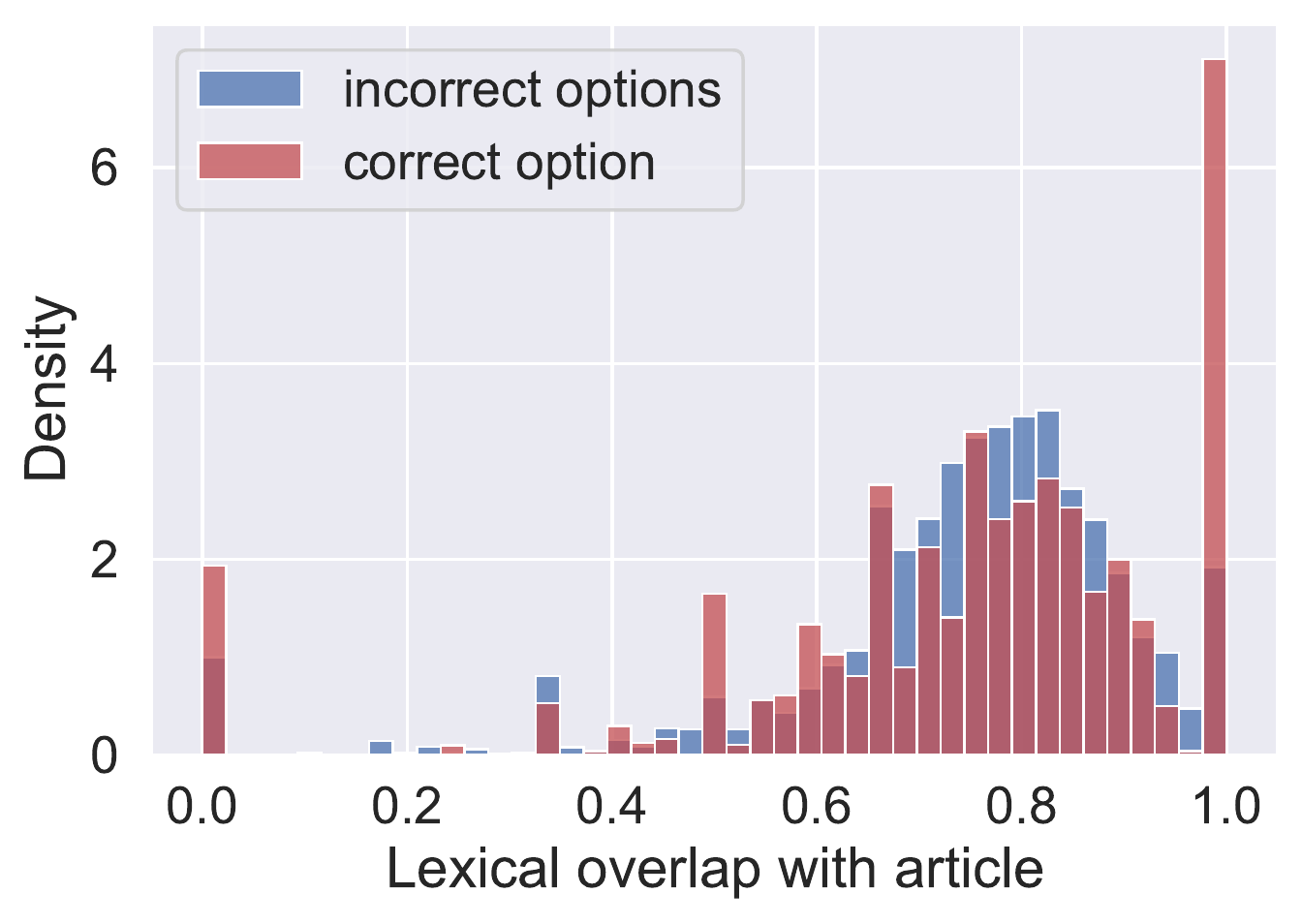}
    \caption{Lexical overlap of all the correct and incorrect options with the article. The distribution is normalized since there are thrice as many incorrect options as there are correct options.}
    \label{fig:lexical_overlap_avg}
\end{figure}

In addition to comparing lexical overlap of the correct option and the maximum lexical overlap of the incorrect option with the article (Section~\ref{ssec:overlap}), we also plot a normalized distribution of lexical overlap for all the correct and incorrect options in Figure~\ref{fig:lexical_overlap_avg}. Despite a higher fraction of the correct options having complete overlap with the article, models would not be able to exploit this heuristic, since other incorrect options for the same question may have complete overlap. This is demonstrated by the plot in Figure~\ref{fig:lexical_overlap_max} and the fact that a baseline which relies on the lexical overlap heuristic only achieves 26.6\% accuracy.

\begin{table*}[th]
\setlength{\tabcolsep}{3.6pt}
\centering
\small
\begin{tabular}{llcccccc}
    \toprule
    & & & \multicolumn{3}{c}{Extraction Based on \textit{Oracle Answers}}  \\
    \cline{4-6} 
    \noalign{\smallskip}
    Training Data & Model & & R-1 & fastText & DPR  \\ 
    \midrule
    QuALITY 
      & RoBERTa-base & & 69.1/67.3 & 61.3/57.2 & 78.3/77.7   \\
      & RoBERTa-large & & 67.5/64.2 & 63.9/60.0 & 75.3/73.7 \\
      & DeBERTaV3-base & & 70.8/68.5 & 65.5/60.8 & 76.4/74.9 \\
      & DeBERTaV3-large & & 68.9/65.2 & 66.6/60.8 & 77.8/75.1 \\
    \midrule 
    RACE 
      & RoBERTa-base & & 54.5/49.9 & 56.1/49.7 & 53.4/47.9  \\
      & RoBERTa-large & & 59.2/53.7 & 58.2/52.1 & 56.9/51.5  \\
      & DeBERTaV3-base & & 56.5/51.5 & 55.9/48.7 & 52.0/45.0 \\
      & DeBERTaV3-large & & 59.8/54.1 & 59.8/53.5 & 57.5/49.6 \\
    \midrule 
    \multirow{3}{2cm}{RACE \\ $\downarrow$ \\ QuALITY}
      & RoBERTa-base & & 67.6/62.8 & 64.6/58.8 & 70.2/67.2 \\
      & RoBERTa-large & & 68.9/63.7 & 66.6/60.5 & 64.1/60.0 \\
      & DeBERTaV3-base & & 69.6/64.4 & 68.1/62.5 & 66.9/61.2 \\
      & DeBERTaV3-large & & 71.0/66.5 & 68.2/62.9 & 71.9/67.1 \\
    \bottomrule
\end{tabular}
\caption{Oracle accuracy on the full QuALITY development set and on the QuALITY-\textsc{hard} subset (full/\textsc{hard}) with models using the \textit{correct answers} as queries to retrieve relevant excerpts. These results are meant to show the relative contribution of the retrieval and reading components of the two-stage models. Caution: These results rely on answers at test time, which are not available to any model during a conventional deployment or test set evaluation, and so are of very limited value in conventional comparisons. 
}
\label{tab:oracle-results}
\end{table*}

\section{More Details on Modeling}
\label{app:modeling}

\subsection{Extraction}

For ROUGE-1 scoring, we use the \texttt{rouge-score} Python package.\footnote{\url{https://github.com/google-research/google-research/tree/master/rouge}}

For fastText scoring, we use SpaCy with the \texttt{en\_core\_web\_sm} model for tokenization, and use embeddings trained on Common Crawl, \footnote{\url{https://dl.fbaipublicfiles.com/fasttext/vectors-english/crawl-300d-2M.vec.zip}} using the top 300k words in the vocabulary.

For DPR, we use the Transformers package \citep{wolf-etal-2020-transformers}, using the \texttt{facebook/} \texttt{dpr-ctx\_encoder-multiset-base} and \texttt{facebook/dpr-question\_encoder-\newline multiset-base} models for encoding the context and query respectively.

\subsection{Training}

\begin{table*}[th]
\setlength{\tabcolsep}{3.0pt}
\centering
\small
\begin{tabular}{lccccc}
    \toprule
    Model & Learning rate & Training time & Max. sequence length & Batch size & Warmup steps \\
    \midrule
    $\text{RoBERTa}_\textsc{Base}$ & 1e-5 & 3160 steps & 512 & 16 & 316 \\
    $\text{RoBERTa}_\textsc{Large}$ & 1e-5 & 3160 steps & 512 & 16 & 316 \\
    $\text{DeBERTaV3}_\textsc{Base}$ & 1e-5 & 3160 steps & 512 & 16 & 316 \\
    $\text{DeBERTaV3}_\textsc{Large}$ & 1e-5 & 3160 steps & 512 & 16 & 316 \\
    $\text{Longformer}_\textsc{Base}$ & 1e-5 & 3160 steps & 4096 & 16 & 316 \\
    $\text{T5}_\textsc{Base}$ & 1e-4 & 40000 steps & 512 & 128 & 0 \\
    % Angie's
    % $\text{LED}_\textsc{Base}$ & 5e-5 & 15 epochs & 16384 & 32 & 1000 \\
    $\text{LED}_\textsc{Base}$ & 1e-5 & 3160 steps & 16384 & 16 & 316 \\
    $\text{LED}_\textsc{Large}$ & 1e-5 & 780 steps & 16384 & 32 & 78 \\
    \bottomrule
\end{tabular}
\caption{Hyperparameters used for fine-tuning models on QuALITY.}
\label{appendix:hyperparamquality}
\end{table*}

\begin{table*}[th]
\setlength{\tabcolsep}{3.0pt}
\centering
\small
\begin{tabular}{lccccc}
    \toprule
    Model & Learning rate & Training time & Max. sequence length & Batch size & Warmup steps \\
    \midrule
    $\text{RoBERTa}_\textsc{Base}$ & 1e-5 & 16473 steps & 512 & 16 & 1647 \\
    $\text{RoBERTa}_\textsc{Large}$ & 1e-5 & 16473 steps & 512 & 16 & 1647 \\
    $\text{DeBERTaV3}_\textsc{Base}$ & 1e-5 & 16473 steps & 512 & 16 & 1647 \\
    $\text{DeBERTaV3}_\textsc{Large}$ & 1e-5 & 16473 steps & 512 & 16 & 1647 \\
    $\text{Longformer}_\textsc{Base}$ & 1e-5 & 16473 steps & 512 & 16 & 1647 \\
    $\text{LED}_\textsc{Base}$ & 1e-5 & 16473 steps & 512 & 16 & 1647 \\
    $\text{LED}_\textsc{Large}$ & 1e-6 & 13727 steps & 512 & 32 & 1372 \\
    \bottomrule
\end{tabular}
\caption{Hyperparameters used for fine-tuning models on RACE.}
\label{appendix:hyperparamrace}
\end{table*}

The full sets of hyperparameters used for tuning our baselines are shown in Table~\ref{appendix:hyperparamquality} and Table~\ref{appendix:hyperparamrace}.

For RoBERTa, DeBERTaV3 and Longformer models, we train on QuALITY for 20 epochs. Where we do intermediate training on RACE, we do so for 3 epochs. Warmup is set to 10\% of the full training steps.

\subsection{Results}
\label{app:results}

Table~\ref{tab:dev-results} shows the results on development set. 
Table~\ref{tab:oracle-results} shows the results using oracle-answer-based extraction. Please refer to the discussion in Section~\ref{sec:results-analysis}.

\end{document}